\documentclass{article}

\usepackage[preprint]{neurips_2023}

\usepackage[utf8]{inputenc} 
\usepackage[T1]{fontenc}    
\usepackage{xcolor}
\definecolor{mydarkblue}{RGB}{0, 2, 115}
\usepackage[colorlinks=true,linkcolor=mydarkblue,citecolor=mydarkblue,urlcolor=mydarkblue,pdfborder={0 0 0}]{hyperref}
\usepackage[toc]{appendix}
\usepackage{url}            
\usepackage{booktabs}       
\usepackage{amsfonts}       
\usepackage{nicefrac}       
\usepackage{microtype}      
\usepackage{xcolor}         
\usepackage{comment}
\usepackage{graphicx}
\usepackage{graphics}
\usepackage{titletoc}
\usepackage{multirow}
\usepackage{wrapfig}
\usepackage{amsmath}
\usepackage{subcaption}
\usepackage{tocloft}
\usepackage{cancel}
\usepackage{adjustbox}
\usepackage{tabularx}
\usepackage{algorithm}
\usepackage{algpseudocode}

\usepackage{fp}
\usepackage{expl3}[2012-07-08]
\ExplSyntaxOn

\ExplSyntaxOff

\usepackage{amsmath,amsfonts,bm}

\def\x{{x}}

\def\xi{{\x_i}}

\newcommand{\ignorethis}[1]{}

\def\eqref#1{equation~\ref{#1}}

\def\1{\bm{1}}

\def\eps{{\epsilon}}

\def\vc{{\bm{c}}}

\def\vf{{\bm{f}}}
\def\vg{{\bm{g}}}

\def\vn{{\bm{n}}}

\def\vx{{\bm{x}}}
\def\vy{{\bm{y}}}

\def\veps{{\bm{\eps}}}

\DeclareMathAlphabet{\mathsfit}{\encodingdefault}{\sfdefault}{m}{sl}
\SetMathAlphabet{\mathsfit}{bold}{\encodingdefault}{\sfdefault}{bx}{n}

\newcommand{\ignore}[1]{}

\makeatletter
\DeclareRobustCommand\onedot{\futurelet\@let@token\@onedot}
\def\@onedot{\ifx\@let@token.\else.\null\fi\xspace}

\makeatother

\definecolor{MyDarkBlue}{rgb}{0,0.08,1}
\definecolor{MyDarkGreen}{rgb}{0.02,0.6,0.02}
\definecolor{MyDarkRed}{rgb}{0.8,0.02,0.02}
\definecolor{MyDarkOrange}{rgb}{0.40,0.2,0.02}
\definecolor{MyPurple}{RGB}{111,0,255}
\definecolor{MyRed}{rgb}{1.0,0.0,0.0}
\definecolor{MyGold}{rgb}{0.75,0.6,0.12}
\definecolor{MyDarkgray}{rgb}{0.66, 0.66, 0.66}
\definecolor{myorange}{RGB}{255,69,0}

\definecolor{revision}{RGB}{255,69,0}

\usepackage[capitalize]{cleveref}
\crefformat{equation}{Eq.~(#2#1#3)}
\crefname{section}{§}{§§}
\Crefname{section}{§}{§§}
\newcommand*\diff{\mathop{}\!\mathrm{d}}

\newcommand{\method}{BOOT}

\newcommand{\emoji}[1]{\raisebox{-0.15\height}{\includegraphics[height=1em]{figures_jpg/#1.png}}}

\titlecontents{paragraph}[5em]{}{}{}{}

\title{{\method}\emoji{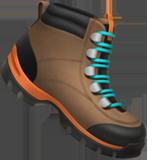}: Data-free Distillation of Denoising Diffusion Models with Bootstrapping}
\author{
Jiatao Gu$^\dagger$, Shuangfei Zhai$^\dagger$, Yizhe Zhang$^\dagger$, Lingjie Liu$^\ddag$, Josh Susskind$^\dagger$\\
$^\dagger$Apple \; $^\ddag$University of Pennsylvania \\
$^\dagger$\texttt{\{jgu32, szhai, yizzhang, jsusskind\}@apple.com} \;
$^\ddag$\texttt{lingjie.liu@seas.upenn.edu}
}

\begin{document}

\maketitle

\begin{abstract}    
Diffusion models have demonstrated excellent potential for generating diverse images. However, their performance often suffers from slow generation due to iterative denoising. Knowledge distillation has been recently proposed as a remedy which can reduce the number of inference steps to one or a few, without significant quality degradation. However, existing distillation methods either require significant amounts of offline computation for generating synthetic training data from the teacher model, or need to perform expensive online learning with the help of real data. In this work, we present a novel technique called \emph{{\method}}, that overcomes these limitations with an efficient data-free distillation algorithm. The core idea is to learn a time-conditioned model that predicts the output of a pre-trained diffusion model teacher given any time-step. Such a model can be efficiently trained based on bootstrapping from two consecutive sampled steps. Furthermore, our method can be easily adapted to large-scale text-to-image diffusion models, which are challenging for conventional methods given the fact that the training sets are often large and difficult to access. We demonstrate the effectiveness of our approach on several benchmark datasets in the DDIM setting, achieving comparable generation quality while being orders of magnitude faster than the diffusion teacher. The text-to-image results show that the proposed approach is able to handle highly complex distributions, shedding light on more efficient generative modeling. Please check our project page: \url{https://jiataogu.me/boot/} for more details.
\begin{figure}[h!]
    \centering
    \vspace{-6pt}
    \includegraphics[width=0.96\linewidth]{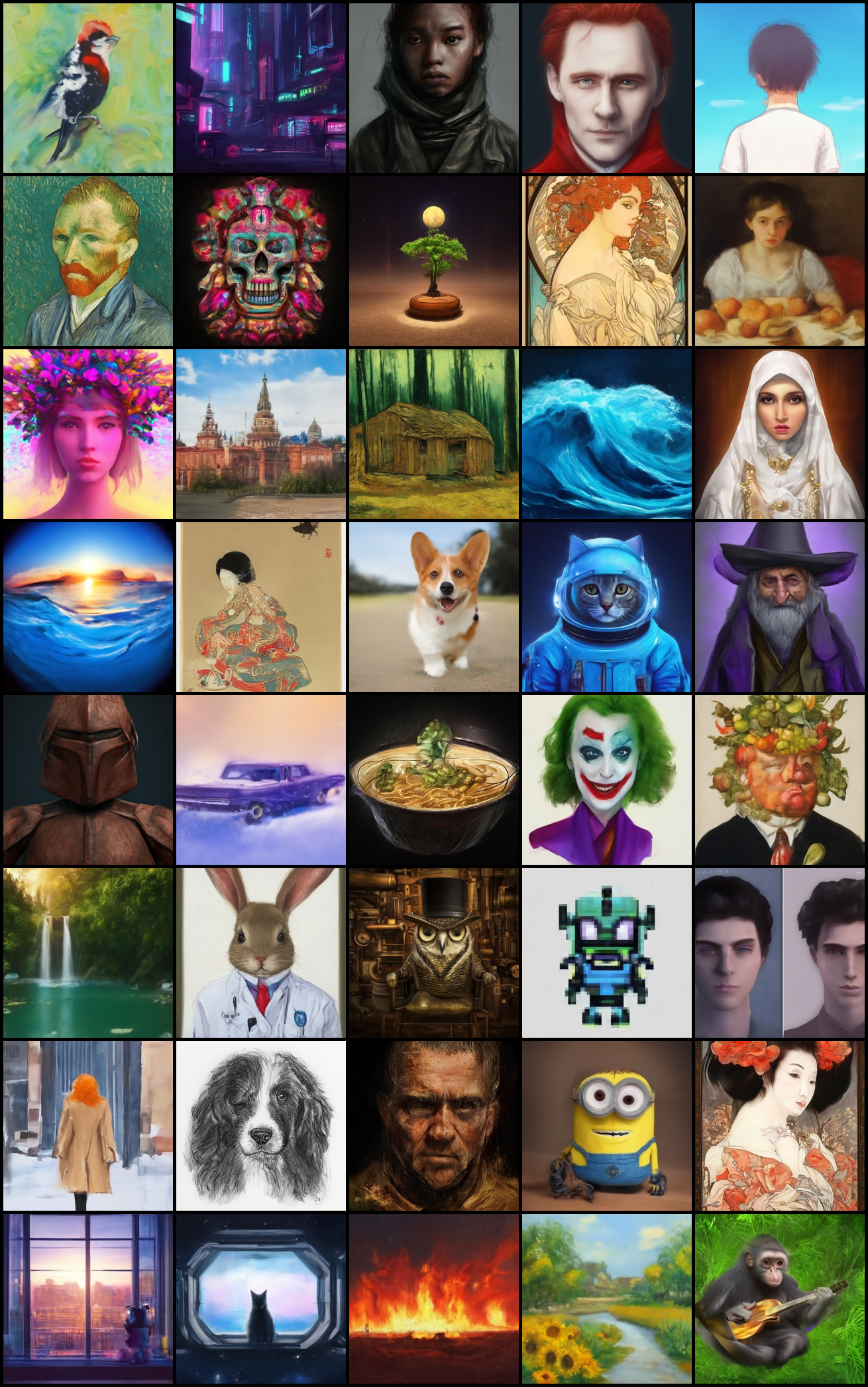}
    \caption{Curated samples of our distilled \textbf{single-step} model with prompts from \emph{diffusiondb}.}
    \label{fig:my_teaser}
\end{figure}
\end{abstract}

\section{Introduction}

\begin{wrapfigure}{r}{0.5\textwidth}
  \centering
  \vspace{-10pt}
  \includegraphics[width=0.49\textwidth]{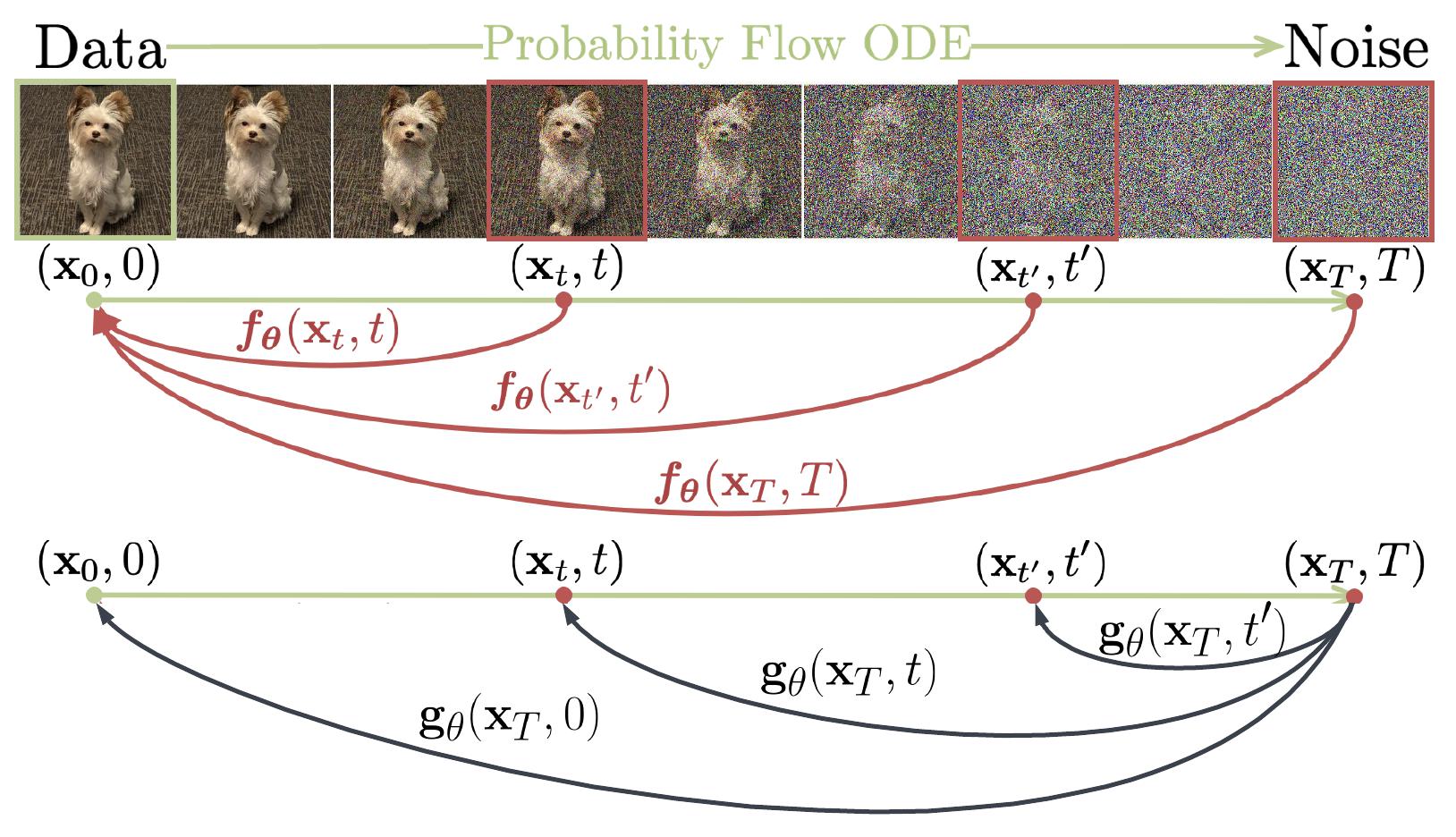}
  \caption{Comparison of Consistency Model~\citep{song2023consistency} ({\textcolor{red}{ red $\uparrow$}}) and {\method} (black $\downarrow$) highlighting the opposing prediction pathways.}
  \vspace{-10pt}
  \label{fig:compare_with_cm}
\end{wrapfigure}

Diffusion models~\citep{sohl2015deep,ho2020denoising,nichol2021improved,song2020score} have become the standard tools for generative applications, such as image~\citep{dhariwal2021diffusion,rombach2021highresolution,ramesh2022hierarchical,saharia2022photorealistic}, video~\citep{ho2022video,ho2022imagen}, 3D~\citep{poole2022dreamfusion,gu2023nerfdiff,liu2023zero1to3,ssdnerf}, audio~\citep{liu2023audioldm}, and text~\citep{li2022diffusion,zhang2023planner} generation. Diffusion models are considered more stable for training compared to alternative approaches like GANs~\citep{Goodfellow14} or VAEs~\citep{kingma2013auto}, as they don't require balancing two modules, making them less susceptible to issues like mode collapse or posterior collapse. Despite their empirical success, standard diffusion models often have slow inference times (around $50\sim1000\times$ slower than single-step models like GANs), which poses challenges for deployment on consumer devices. This is mainly because diffusion models use an iterative refinement process to generate samples.

To address this issue, previous studies have proposed using knowledge distillation to improve the inference speed~\citep{hinton2015distilling}. The idea is to train a faster student model that can replicate the output of a pre-trained diffusion model. In this work, we focus on learning single-step models that only require one neural function evaluation (NFE). However, conventional methods, such as \citet{luhman2021knowledge}, require executing the full teacher sampling to generate synthetic targets for every student update, which is impractical for distilling large diffusion models like StableDiffusion~\citep[SD,][]{rombach2021highresolution}.
Recently, several techniques have been proposed to avoid sampling using the concept of "bootstrap". For example, \citet{salimans2022progressive} gradually reduces the number of inference steps based on the previous stage's student, while \citet{song2023consistency} and \citet{berthelot2023tract} train single-step denoisers by enforcing self-consistency between adjacent student outputs along the same diffusion trajectory (see \cref{fig:compare_with_cm}). However, these approaches rely on the availability of real data to simulate the intermediate diffusion states as input, which limits their applicability in scenarios where the desired real data is not accessible.

In this paper, we propose \emph{\method}, a data-free knowledge distillation method for denoising diffusion models based on bootstrapping. {\method} is partially motivated by the observation made by consistency model~\citep[CM,][]{song2023consistency} that all points on the same diffusion trajectory (also known as PF-ODE~\citep{song2020score}) have a deterministic mapping between each other. Unlike CM, which seeks self-consistency from any $\vx_t$ to $\vx_0$, {\method} predicts all possible $\vx_t$ given the same noise point $\veps$ and a time indicator $t$. Since our model $\vg_\theta$ always reads pure Gaussian noise, there is no need to sample from real data. Moreover, learning all $\vx_t$ from the same $\veps$ enables bootstrapping: it is easier to predict $\vx_t$ if the model has already learned to generate $\vx_{t'}$ where $t' > t$. However, formulating bootstrapping in this way presents additional challenges, such as noisy sample prediction, which is non-trivial for neural networks. To address this, we learn the student model from a novel \emph{Signal-ODE} derived from the original PF-ODE. We also design objectives and boundary conditions to enhance the sampling quality and diversity. This enables efficient inference of large diffusion models in scenarios where the original training corpus is inaccessible due to privacy or other concerns. For example, we can obtain an efficient model for synthesizing images of \emph{"raccoon astronaut"} by distilling the text-to-image model with the corresponding prompts (shown in \cref{fig:pipeline}), even though collecting such data in reality is difficult.

In the experiments, we first demonstrate the efficacy of {\method} on various challenging image generation benchmarks, including unconditional and class-conditional settings.
Next, we show that the proposed method can be easily adopted to distill text-to-image diffusion models.
An illustration of sampled images from our distilled text-to-image model is shown in \cref{fig:my_teaser}.

\section{Preliminaries}
\subsection{Diffusion Models}
Diffusion models~\citep{sohl2015deep,song2019generative,ho2020denoising} belong to a class of deep generative models that generate data by progressively removing noise from the initial input. In this work, we focus on continuous-time diffusion models~\citep{song2020score,kingma2021variational,karras2022elucidating} in the variance-preserving formulation~\citep{salimans2022progressive}. Given a data point $\vx \in \mathbb{R}^N$, we model a series of time-dependent latent variables $\{\vx_t | t\in [0, T], \vx_0=\vx\}$ based on a given noise schedule $\{{\alpha_t, \sigma_t}\}$:
\begin{equation*}
    q(\vx_t | \vx_s) = \mathcal{N}(\vx_t; \alpha_{t|s}\vx_s, \sigma^2_{t|s}I), \ \ \text{and} \ \ q(\vx_t | \vx) = \mathcal{N}(\vx_t;\alpha_t\vx, \sigma_t^2I),
\end{equation*}
where $\alpha_{t|s} = \alpha_t/\alpha_s$ and $\sigma^2_{t|s}=\sigma_t^2-\alpha_{t|s}^2\sigma_s^2$ for $s < t$. By default, the signal-to-noise ratio (SNR, ${\alpha_t^2}/{\sigma_t^2}$) decreases monotonically with $t$. A diffusion model $\vf_\phi$ learns to reverse the diffusion process by denoising $\vx_t$, which can be easily sampled given the real data $\vx$ with $q(\vx_t|\vx)$:
\begin{equation}
    \mathcal{L}^\textrm{Diff}_\phi = \mathbb{E}_{\vx_t\sim q(\vx_t | \vx), t \sim [0, T]}\left[\omega_t\cdot\|\vf_\phi(\vx_t, t) - \vx\|_2^2\right].
    \label{eq.learn_dpm}
\end{equation}
Here, $\omega_t$ is the weight used to balance perceptual quality and diversity. The parameterization of $\phi$ typically involves U-Net~\citep{ronneberger2015u,dhariwal2021diffusion} or Transformer~\citep{peebles2022scalable,bao2022all}. In this paper, we use $\vf_\phi$ to represent signal predictions. However, due to the mathematical equivalence of signal, noise, and $\textrm{v}$-predictions~\citep{salimans2022progressive} in the denoising formulation, the loss function can also be defined based on noise or $\textrm{v}$-predictions. For simplicity, we use $\vf_\phi$ for all cases in the remainder of the paper.

One can use ancestral sampling~\citep{ho2020denoising} to synthesize new data from the learned model. While the conventional method is stochastic, DDIM~\citep{song2020denoising} demonstrates that one can follow a deterministic  sampler to generate the final sample $\vx_0$, which follows the update rule:
\begin{equation}
    \vx_s = \left({\sigma_s}/{\sigma_t}\right)\vx_t + \left(\alpha_s - \alpha_t{\sigma_s}/{\sigma_t}\right)\vf_\phi(\vx_t, t),\ \ \ s < t, 
    \label{eq.ddim}
\end{equation}
with the boundary condition $\vx_T = \veps \sim \mathcal{N}(0, I)$. As noted in \citet{lu2022dpm}, \cref{eq.ddim} is equivalent to the first-order ODE solver for the underlying probability-flow (PF) ODE~\citep{song2020score}. Therefore, the step size $\delta=t-s$ needs to be small to mitigate error accumulation. Additionally, using higher-order solvers such as Runge-Kutta~\citep{suli2003introduction}, Heun~\citep{ascher1998computer}, and other solvers~\citep{lu2022dpm,jolicoeur2021gotta} can further reduce the number of function evaluations (NFEs). However, these approaches are not applicable in single-step.

\subsection{Knowledge Distillation}
Orthogonal to the development of ODE solvers, distillation-based techniques have been proposed to learn faster student models from a pre-trained diffusion teacher. The most straightforward approach is to perform \textbf{direct distillation}~\citep{luhman2021knowledge}, where a student model $\vg_\theta$ is trained to learn from the output of the diffusion model, which is computationally expensive itself:
\begin{equation}
    \mathcal{L}^\textrm{Direct}_\theta = \mathbb{E}_{\veps\sim \mathcal{N}(0, I)} \| \vg_\theta(\veps) - \texttt{ODE-Solver}(\vf_\phi, \veps, T\rightarrow 0)\|^2_2,
\end{equation}
Here, $\texttt{ODE-solver}$ refers to any solvers like DDIM as mentioned above. While this naive approach shows promising results, it typically requires over 50 steps of evaluations to obtain reasonable distillation targets, which becomes a bottleneck when learning large-scale models.

Alternatively, recent studies~\citep{salimans2022progressive,song2023consistency,berthelot2023tract} have proposed methods to avoid running the full diffusion path during distillation. For instance, the consistency model~\citep[CM,][]{song2023consistency} trains a time-conditioned student model $\vg_\theta(\vx_t, t)$ to predict self-consistent outputs along the diffusion trajectory in a bootstrap fashion:
\begin{equation}
    \mathcal{L}^\textrm{CM}_\theta = \mathbb{E}_{\vx_t\sim q(\vx_t|\vx), s,t\sim[0,T], s<t} \|\vg_\theta(\vx_t, t) - \vg_{\theta^-}({\vx}_s, s) \|^2_2,
    \label{eq.cm_loss}
\end{equation}
where ${\vx}_s = \texttt{ODE-Solver}(\vf_\phi, \vx_t, t\rightarrow s)$, typically with a single-step evaluation using \cref{eq.ddim}. In this case, $\theta^-$ represents an exponential moving average (EMA) of the student parameters $\theta$, which is important to prevent the self-consistency objectives from collapsing into trivial solutions by always predicting similar outputs. After training, samples can be generated by executing $\vg_\theta(\vx_T, T)$ with a single NFE.
It is worth noting that \cref{eq.cm_loss} requires sampling $\vx_t$ from the real data sample $\vx$, which is the essence of bootstrapping: the model learns to denoise increasingly noisy inputs until $\vx_T$. However, in many tasks, the original training data $\vx$ for distillation is inaccessible. For example, text-to-image generation models require billions of paired data for training. One possible solution is to use a different dataset for distillation; however, the mismatch in the distributions of the two datasets would result in suboptimal distillation performance.

\section{Method}
\begin{figure}[t]
    \centering
    \includegraphics[width=\linewidth]{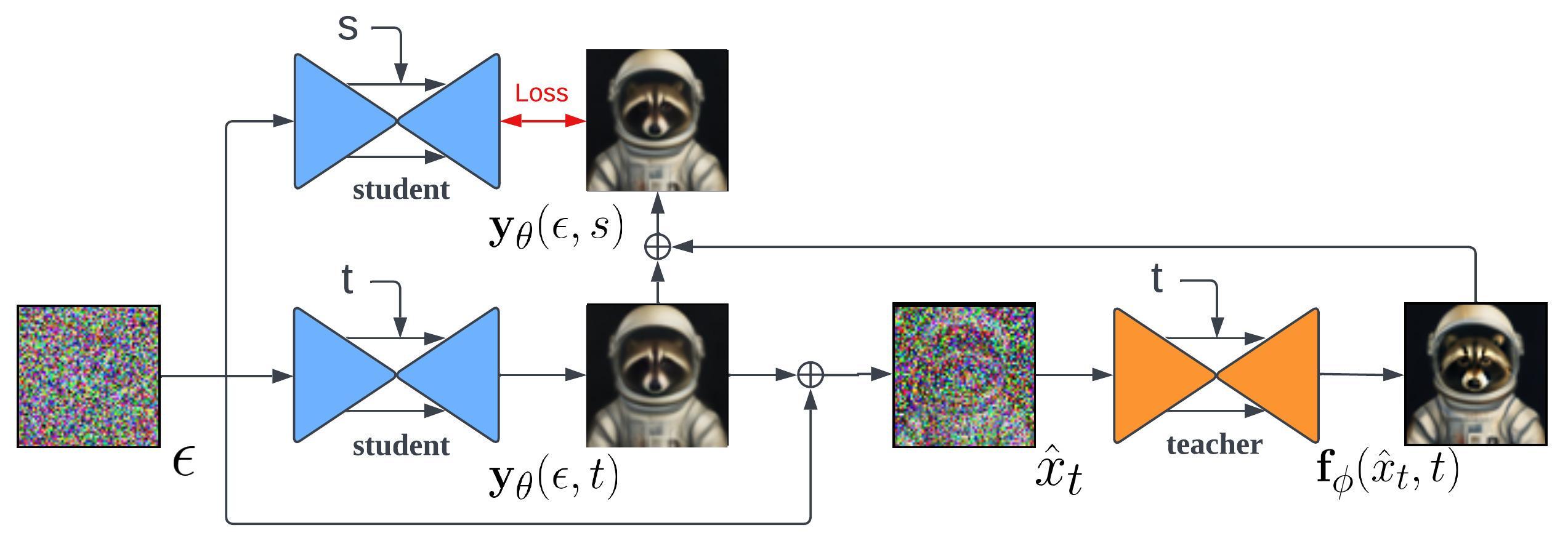}
    \caption{Training pipeline of {\method}.
    $s$ and $t$ are two consecutive timesteps where $s<t$.
    From a noise map $\veps$, the objective of {\method} minimizes the difference between the output of a student model at timestep $s$, and the output of stacking the same student model and a teacher model at an earlier time $t$. {\bf The whole process is data-free}. }
    \vspace{-10pt}
    \label{fig:pipeline}
\end{figure}

In this section, we present {\method}, a novel distillation approach inspired by the concept of bootstrapping without requiring target domain data during training. We begin by introducing \emph{signal-ODE}, a modeling technique focused exclusively on signals (\cref{sec.signal}), and its corresponding distillation process (\cref{sec.learning}). Subsequently, we explore the application of {\method} in text-to-image generation (\cref{sec.t2i}). The training pipeline is depicted in \cref{fig:pipeline}, providing an overview of the process.
\subsection{Signal-ODE}
\label{sec.signal}
We utilize a time-conditioned student model $\vg_\theta(\veps, t)$ in our approach. Similar to direct distillation~\citep{luhman2021knowledge}, {\method} always takes random noise $\veps$ as input and approximates the intermediate diffusion model variable: $\vg_\theta(\veps, t)\approx \vx_t = \texttt{ODE-Solver}(\vf_\phi, \veps, T\rightarrow t), \veps \sim \mathcal{N}(0, I)$. This approach eliminates the need to sample from real data during training. The final sample can be obtained as $\vg_\theta(\veps, 0)\approx\vx_0$. However, it poses a challenge to train $\vg_\theta$ effectively, as neural networks struggle to predict partially noisy images~\citep{berthelot2023tract}, leading to out-of-distribution (OOD) problems and additional complexities in learning $\vg_\theta$ accurately.

To overcome the aforementioned challenge, we propose an alternative approach where we predict $\vy_t = (\vx_t - \sigma_t \veps) / \alpha_t$. In this case, $\vy_t$ represents the low-frequency "signal" component of $\vx_t$, which is easier for neural networks to learn. The initial noise for diffusion is denoted by $\veps$. This prediction target is reasonable since it aligns with the boundary condition of the teacher model, where $\vy_0 = \vx_0$. Furthermore, we can derive an iterative equation from \cref{eq.ddim} for consecutive timesteps:
\begin{equation}
     \vy_s = \left(1 - e^{\lambda_s - \lambda_t}\right) \vf_\phi(\vx_t, t) + e^{\lambda_s - \lambda_t}\vy_t,
    \label{eq.s_iter}
\end{equation}
where $\vx_t=\alpha_t\vy_t+\sigma_t\veps$, and $\lambda_t = -\log(\alpha_t / \sigma_t)$ represents the "negative half log-SNR." Notably, the noise term $\veps$ automatically cancels out in \cref{eq.s_iter}, indicating that the model always learns from the signal space. Moreover, \cref{eq.s_iter} demonstrates an interpolation between the current model prediction and the diffusion-denoised output.
Similar to the connection between DDIM and PF-ODE~\citep{song2020score}, we can also obtain a continuous version of \cref{eq.s_iter} by letting $s \rightarrow t^-$ as follows: 
\begin{equation}
    \frac{\diff\vy_t}{\diff t} = -\lambda'_t\cdot\left(\vf_\phi(\vx_t, t) - \vy_t \right), \;\;\; \vy_{T} \sim p_\veps
    \label{eq.s_ode}
\end{equation}
where $\lambda'_t={\diff{\lambda}}/{\diff{t}}$, and $p_\veps$ epresents the boundary distribution of $\vy_t$. It's important to note that \cref{eq.s_ode} differs from the PF-ODE, which directly relates to the score function of the data. In our case, the ODE, which we refer to as "Signal-ODE," is specifically defined for signal prediction. At each timestep $t$, a fixed noise $\veps$ is injected and denoised by the diffusion model $\vf_\phi$. The Signal-ODE implies a "ground-truth" trajectory for sampling new data. For example, one can initialize a reasonable $\vy_T=\veps\sim\mathcal{N}(0, I)$ and solve the Signal-ODE to obtain the final output $\vy_0$. Although the computational complexity remains the same as conventional DDIM, we will demonstrate in the next section how we can efficiently approximate $\vy_t$ using bootstrapping objectives.

\subsection{Learning with Bootstrapping}
\label{sec.learning}
Our objective is to learn $\vy_\theta(\veps,t)\approx\vy_t$  as a single-step prediction model using neural networks, rather than solving the signal-ODE with \cref{eq.s_ode}. By matching both sides of \cref{eq.s_ode}, we can readily obtain the loss function:
\begin{equation}
     \mathcal{L}^\textrm{DE}_\theta = \mathbb{E}_{\veps\sim \mathcal{N}(0, I), t\sim [0, T]}
     \left|\left|
        \frac{\diff\vy_\theta(\veps, t)}{\diff t} + \lambda'_t\cdot\left(\vf_\phi(\hat{\vx}_t, t) -\vy_\theta(\veps, t) \right)
     \right|\right|^2_2.
     \label{eq.de}
\end{equation}
In \cref{eq.de}, we use $\vy_\theta(\veps, t)$ to estimate $\vy_t$, and $\hat{\vx}_t= \alpha_t\vy_\theta(\veps, t) + \sigma_t\veps$ represents the corresponding noisy image. Instead of using forward-mode auto-differentiation, which can be computationally expensive, we can approximate the above equation with finite differences due to the 1-dimensional nature of $t$. The approximate form is similar to \cref{eq.s_iter}:
\begin{equation}
     \mathcal{L}^\textrm{BS}_\theta = \mathbb{E}_{\veps\sim \mathcal{N}(0, I), t\sim [\delta, T]}
    \left[
    \frac{\tilde{w}_t}{\delta^2}
    \left|\left|\vy_\theta(\veps, s) - 
    \texttt{SG}\left[\vy_\theta(\veps, t) +
    \underbrace{\delta\lambda'_t\cdot\left(
        \left(\vf_\phi(\hat{\vx}_t, t)\right) - \vy_\theta(\veps, t) 
    \right)}_{\textrm{incremental improvement}}
    \right]
    \right|\right|^2_2
    \right],
    \label{eq.training}
\end{equation}
where $s = t - \delta$ and $\delta$ is the discrete step size. $\tilde{w}_t$ represents the time-dependent loss weighting, which can be chosen uniformly. We use $\texttt{SG}[.]$ as the stop-gradient operator for training stability. 

Unlike CM-based methods, such as those mentioned in \cref{eq.cm_loss}, we do not require an exponential moving average (EMA) copy of the student parameters to avoid collapsing. This avoids potential slow convergence and sub-optimal solutions. As shown in \cref{eq.training}, the proposed objective is unlikely to degenerate because there is an incremental improvement term in the training target, which is mostly non-zero. In other words, we can consider $\vy_\theta$ as an exponential moving average of $\vf_\phi$, with a decaying rate of $1 - \delta\lambda'_t$. This ensures that the student model always receives distinguishable signals for different values of $t$.

\vspace{-5pt}\paragraph{Error Accumulation}
A critical challenge in learning {\method} is the "error accumulation" issue, where imperfect predictions of $\vy_\theta$ on large $t$ can propagate to subsequent timesteps. While similar challenges exist in other bootstrapping-based approaches, it becomes more pronounced in our case due to the possibility of out-of-distribution inputs $\hat{\vx}_t$ for the teacher model, resulting from error accumulation and leading to incorrect learning signals. To mitigate this, we employ two methods:
(1) We uniformly sample $t$ throughout the training time, despite the potential slowdown in convergence.
(2) We use a higher-order solver (e.g., Heun's method~\citep{ascher1998computer}) to compute the bootstrapping target with better estimation.
\begin{figure}[t]
    \centering
    \includegraphics[width=0.97\linewidth]{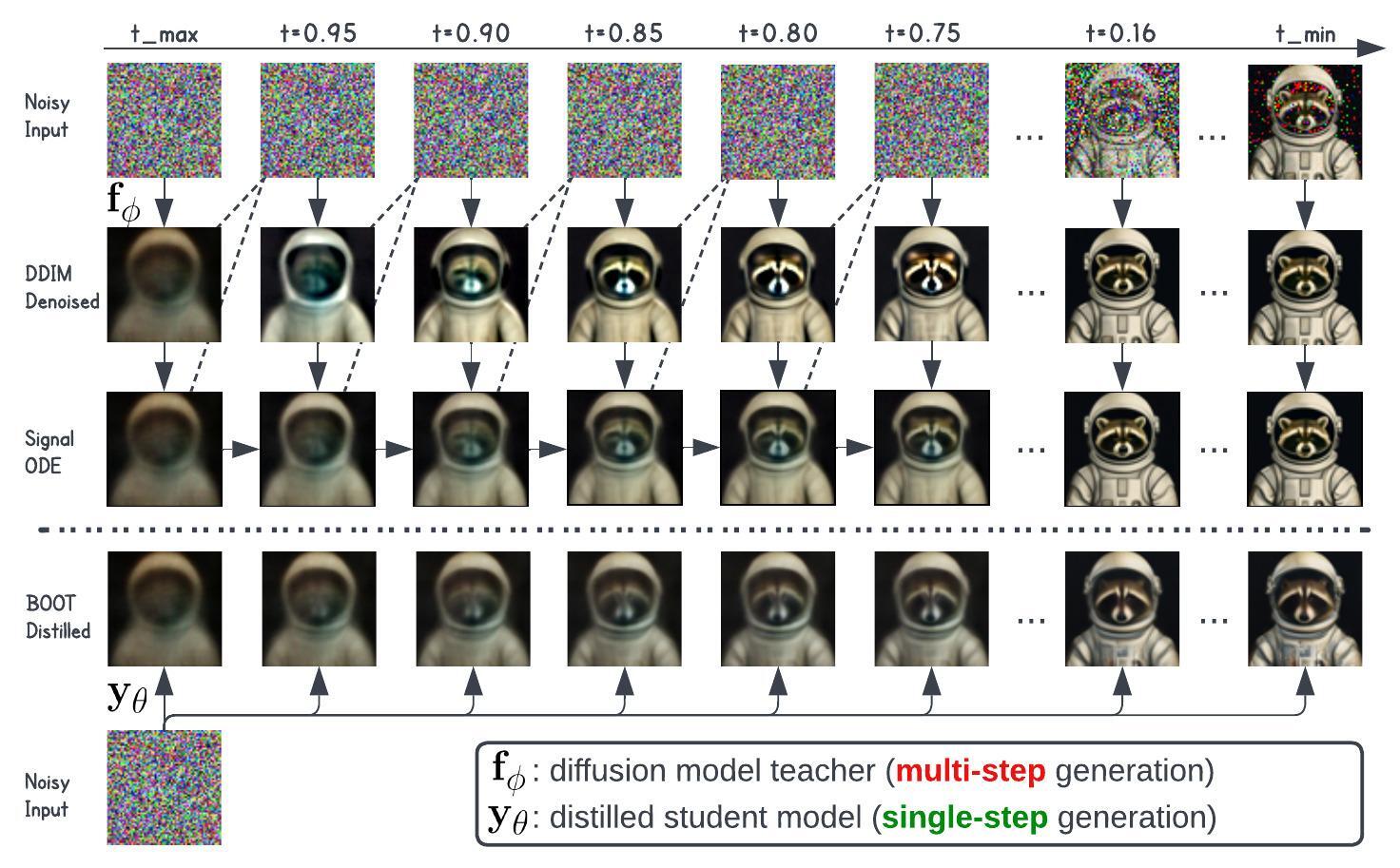}
    \caption{Comparison between the generated outputs of DDIM/Signal-ODE and our distilled model given the same prompt \emph{A raccoon wearing a space suit, wearing a helmet. Oil painting in the style of Rembrandt} and initial noise input. 
    By definition, signal-ODE converges to the same final sample as the original DDIM, while the distilled single-step model does not necessarily follow. 
    }
    \vspace{-5pt}
    \label{fig:visualization}
\end{figure}

\vspace{-5pt}\paragraph{Boundary Condition}
In theory, the boundary $\vy_T$ can have arbitrary values since $\alpha_T=0$, and the value of $\vy_T$ does not affect the value $\vx_T=\veps$. However, $\lambda'_t$ is unbounded at $t=T$, leading to numerical issues in optimization. As a result, the student model must be learned within a truncated range $t \in [t_{\min}, t_{\max}]$. This necessitates additional constraints at the boundaries to ensure that $\alpha_{t_{\max}}\vy_\theta(\veps, t_{\max}) + \sigma_{t_{\max}}\veps$ follows the same distribution as the diffusion model. In this work, we address this through an auxiliary boundary loss:
\begin{equation}
    \mathcal{L}_\theta^\textrm{BC} = \mathbb{E}_{\veps\sim\mathcal{N}(0,I)} \left[\| 
    \vf_\phi(\veps, t_{\max}) - \vy_\theta(\veps, t_{\max}) 
    \|^2_2\right].
    \label{eq.boundary}
\end{equation}
Here, we enforce the student model to match the initial denoising output. In our early exploration, we found that the boundary condition is crucial for the single-step student to fully capture the modeling space of the teacher, especially in text-to-image scenarios. Failure to learn the boundaries tends to result in severe mode collapse and color-saturation problems.

The overall learning objective combines $\mathcal{L}_\theta = \mathcal{L}^\textrm{BS}_\theta + \beta \mathcal{L}^\textrm{BC}_\theta$, where $\beta$ is a hyper-parameter. The algorithm for student model distillation is presented in Appendix~\cref{alg:distillation}.

\subsection{Distillation of Text-to-Image Models}
\label{sec.t2i}
\paragraph{Distillation with Guidance}
Our approach can be readily applied for distilling conditional diffusion models, such as text-to-image generation~\citep{ramesh2022hierarchical,rombach2021highresolution,balaji2022ediffi}, where a conditional denoiser $\vf_\phi(\vx_t, t, \vc)$ is learned with the same objective given an aligned dataset. 
In practice, inference of these models requires necessary post-processing steps for augmenting the conditional generation. For instance, one can perform classifier-free guidance~\citep[CFG,][]{ho2022classifier} to amplify the conditioning:
\begin{equation}
	\tilde{\vf_\phi}(\vx_t, t, \vc) = \vf_\phi(\vx_t, t, \vn) + w \cdot \left(\vf_\phi(\vx_t, t, \vc)  - \vf_\phi(\vx_t, t, \vn) \right), 
	\label{eq.cfg}
\end{equation}
where $\vn$ is the negative prompt (or empty), and $w$ is the guidance weight (by default $w=7.5$) over the denoised signals. We directly use the modified $\tilde{\vf_\phi}$ to replace the original $\vf_\phi$ in the training objectives in ~\cref{eq.training,eq.boundary}. Optionally, similar to~\citet{meng2022distillation}, we can also learn student model condition on both $t$ and $w$ to reflect different guidance strength. 

\vspace{-5pt}\paragraph{Pixel or Latent}
Our method can be easily adopted in either pixel~\citep{saharia2022photorealistic} or latent space~\citep{rombach2021highresolution} models without specific code change. 
 For pixel-space models, it is sometimes critical to apply clipping or dynamic thresholding~\citep{saharia2022photorealistic} over the denoised targets to avoid over-saturation. Similarly, we also clip the targets in our objectives \cref{eq.training,eq.boundary}.
Pixel-space models~\citep{saharia2022photorealistic} typically involve learning cascaded models (one base model + a few super-resolution (SR) models) to increase the output resolutions progressively.
We can also distill the SR models with {\method} into one step by conditioning both the SR teacher and the student with the output of the distilled base model.

\section{Experiments}
\subsection{Experimental Setups}
\begin{figure}[t]
  \centering
  \vspace{-7pt}  \includegraphics[width=\textwidth]{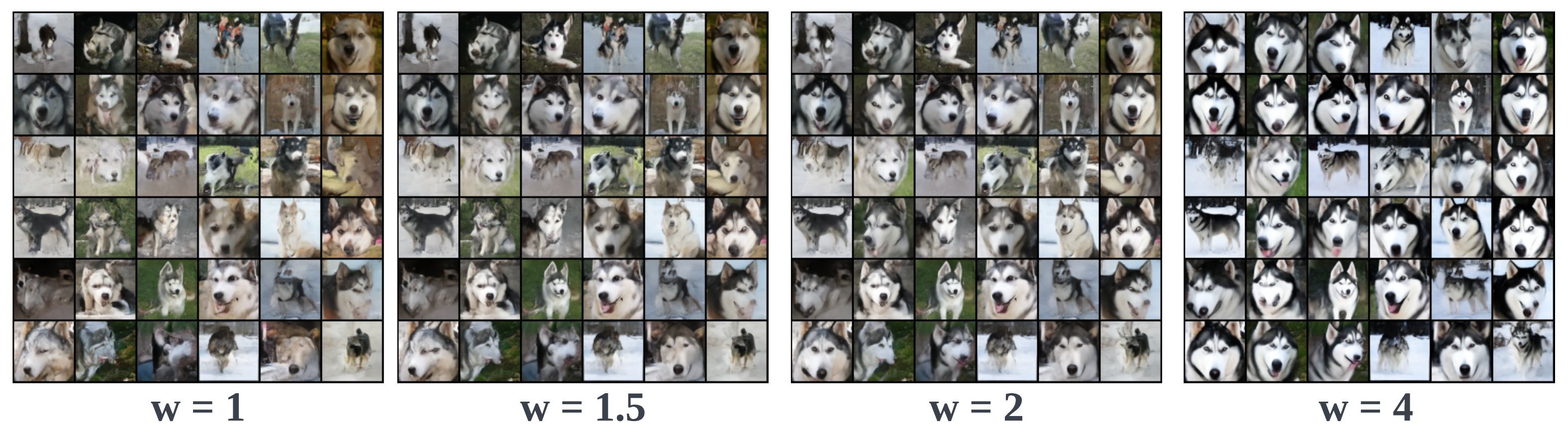}
  \caption{The distilled student is able to trade generation quality with diversity based on CFG weights.}
    \vspace{-12pt}
  \label{fig:cfg_imagenet}

\end{figure}

\paragraph{Diffusion Model Teachers}
We begin by evaluating the performance of {\method} on diffusion models trained on standard image generation benchmarks:  FFHQ $64\times 64$~\citep{karras2017progressive}, class-conditional ImageNet $64 \times 64$~\citep{Deng2009ImageNet:Database} and LSUN Bedroom $256\times 256$~\citep{yu2015lsun}.
To ensure a fair comparison, we train all teacher diffusion models separately on each dataset using the signal prediction objective. Additionally, for ImageNet, we test the performance of CFG where the student models are trained with random conditioning on $w \in [1, 5]$ (see the effects of $w$ in \cref{fig:cfg_imagenet}).
\begin{table}[t]
    \centering
    \small
    \begin{tabular}{lc cc cc cc}
    \toprule
        & 
        \multirow{2}{*}{Steps} & 
        \multicolumn{2}{c}{\bf FFHQ $\mathbf{64\times 64}$} & 
        \multicolumn{2}{c}{\bf LSUN $\mathbf{256\times 256}$} & 
        \multicolumn{2}{c}{\bf ImageNet $\mathbf{64\times 64}$} \\
        & & {\bf FID / Prec. / Rec.} & {\bf fps} & {\bf FID / Prec. / Rec.} & {\bf fps} & {\bf FID / Prec. / Rec.} & {\bf fps} \\
         \midrule 
        DDPM & 250 & \;\;5.4 / 0.80 / 0.54 & 0.2 & 8.2 / 0.55 / 0.43 & 0.1 &11.0 / 0.67 / 0.58 & 0.1 \\
        \midrule
        \multirow{3}{*}{DDIM} 
        &  50 & \;\;7.6 / 0.79 / 0.48 & 1.2 & 13.5 / 0.47 / 0.40 & 0.6 & 13.7 / 0.65 / 0.56 & 0.6 \\ 
        &  10 & 18.3 / 0.78 / 0.27 & 5.3 & 31.0 / 0.27 / 0.32 & 3.1 & 18.3 / 0.60 / 0.49 & 3.3 \\
        &  1 & 225 / 0.10 / 0.00 &  54 & 308 / 0.00 / 0.00 & 31 & 237 / 0.05 / 0.00 & 34\\
        \midrule
        Ours & 1 & \;\;9.0 / 0.79 / 0.38 &  54 &  23.4 / 0.38 / 0.29 & 32 & 16.3 / 0.68 / 0.36 & 34\\ 
    \bottomrule
    \end{tabular}
    \vspace{5pt}
    \caption{Comparison for image generation benchmarks on FFHQ, LSUN and class-conditioned ImageNet. For ImageNet, numbers are reported without using CFG ($w=1$).}
    \vspace{-15pt}
    \label{tab:ffhq_imagenet}
\end{table}
\begin{figure}[t]
    \centering
    \includegraphics[width=\linewidth]{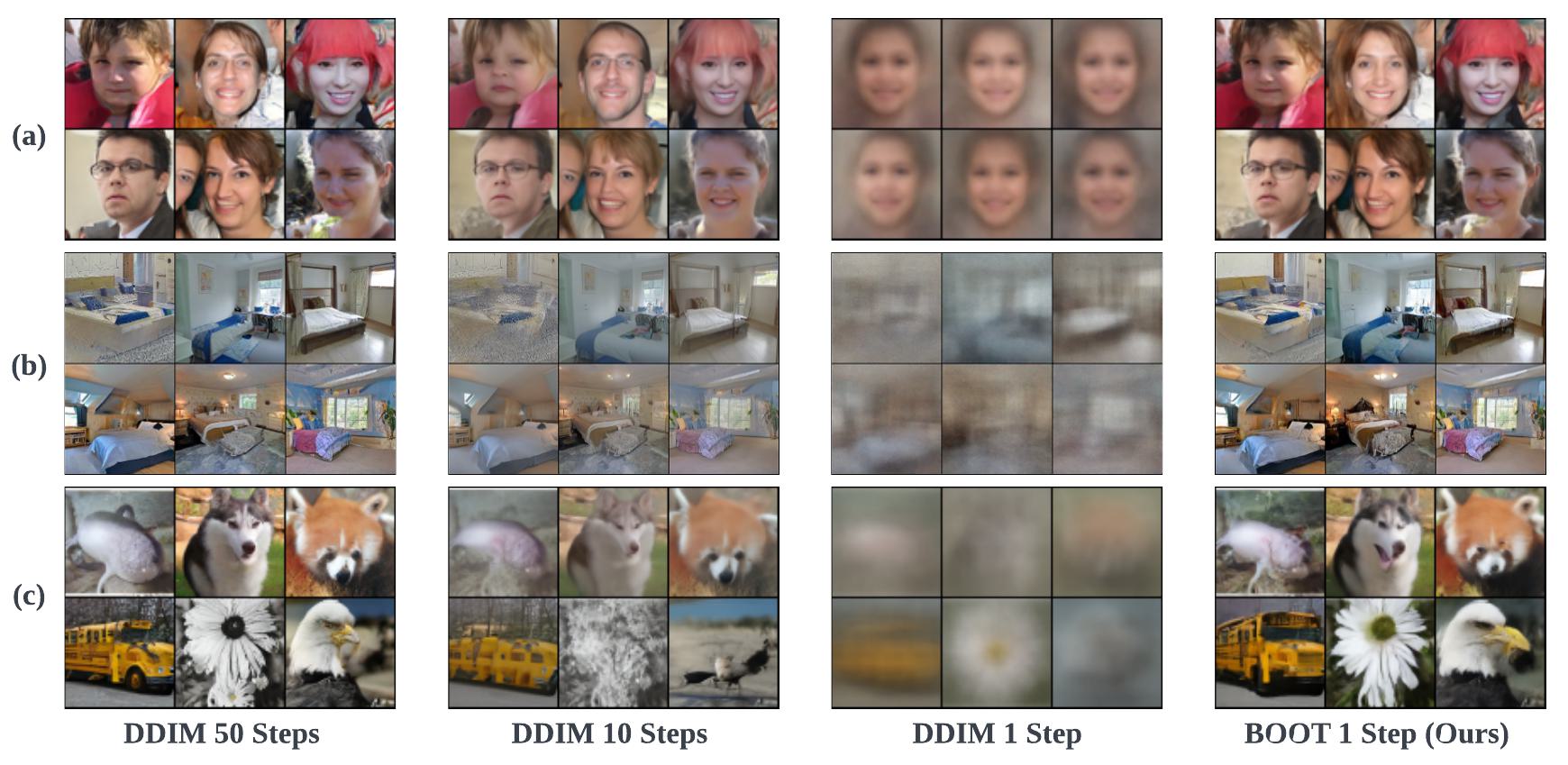}
    \caption{Uncurated samples of  \{50, 10, 1\} DDIM sampling steps and the proposed {\method} from (a) FFHQ (b) LSUN (c) ImageNet benchmarks, respectively, given the same set of initial noise input.}
    \label{fig:image}
    \vspace{-7pt}
\end{figure}

For text-to-image generation scenarios, we directly apply {\method} on open-sourced diffusion models in both
pixel-space~\citep[DeepFloyd-IF (IF),][]{saharia2022photorealistic} \footnote{\url{https://github.com/deep-floyd/IF}}
and latents space~\citep[StableDiffusion (SD),][]{rombach2021highresolution}
\footnote{\url{https://github.com/Stability-AI/stablediffusion}}. 
Thanks to the data-free nature of {\method}, we do not require access to the original training set, which may consist of billions of text-image pairs with unknown preprocessing steps. Instead, we only need the prompt conditions to distill both models. In this work, we consider general-purpose prompts generated by users. Specifically, we utilize diffusiondb~\citep{wangDiffusionDBLargescalePrompt2022}, a large-scale prompt dataset that contains $14$ million images generated by StableDiffusion using prompts provided by real users. We only utilize the text prompts for distillation.

\vspace{-5pt}\paragraph{Implementation Details}
Similar to previous research~\citep{song2023consistency}, we use student models with architectures similar to those of the teachers, having nearly identical numbers of parameters. A more comprehensive architecture search is left for future work. We initialize the majority of the student $\vy_\theta$ parameters with the teacher model $\vf_\phi$, except for the newly introduced conditioning modules (target timestep $t$ and potentially the CFG weight $w$), which are incorporated into the U-Net architecture in a similar manner as how class labels were incorporated. It is important to note that the target timestep $t$ is different from the original timestep used for conditioning the diffusion model, which is always set to $t_{\max}$ for the student model. Based on the actual implementation of the teacher models, we initialize the student output accordingly to accommodate the pretrained weights:
    $\vy_\theta(\veps) = (\texttt{NN}_x(\veps)) \vee (\veps - \texttt{NN}_\epsilon(\veps)) \vee
(-\texttt{NN}_v(\veps))$,
where $\vee$ represents ``or'' and $\texttt{NN}_x, \texttt{NN}_\epsilon, \texttt{NN}_v$ correspond to the pre-trained teacher networks using the signal, noise or velocity~\citep{salimans2022progressive} parameterization, respectively. We include additional details in the \cref{app:details}.

\begin{figure}[t]
    \centering\includegraphics[width=\linewidth]{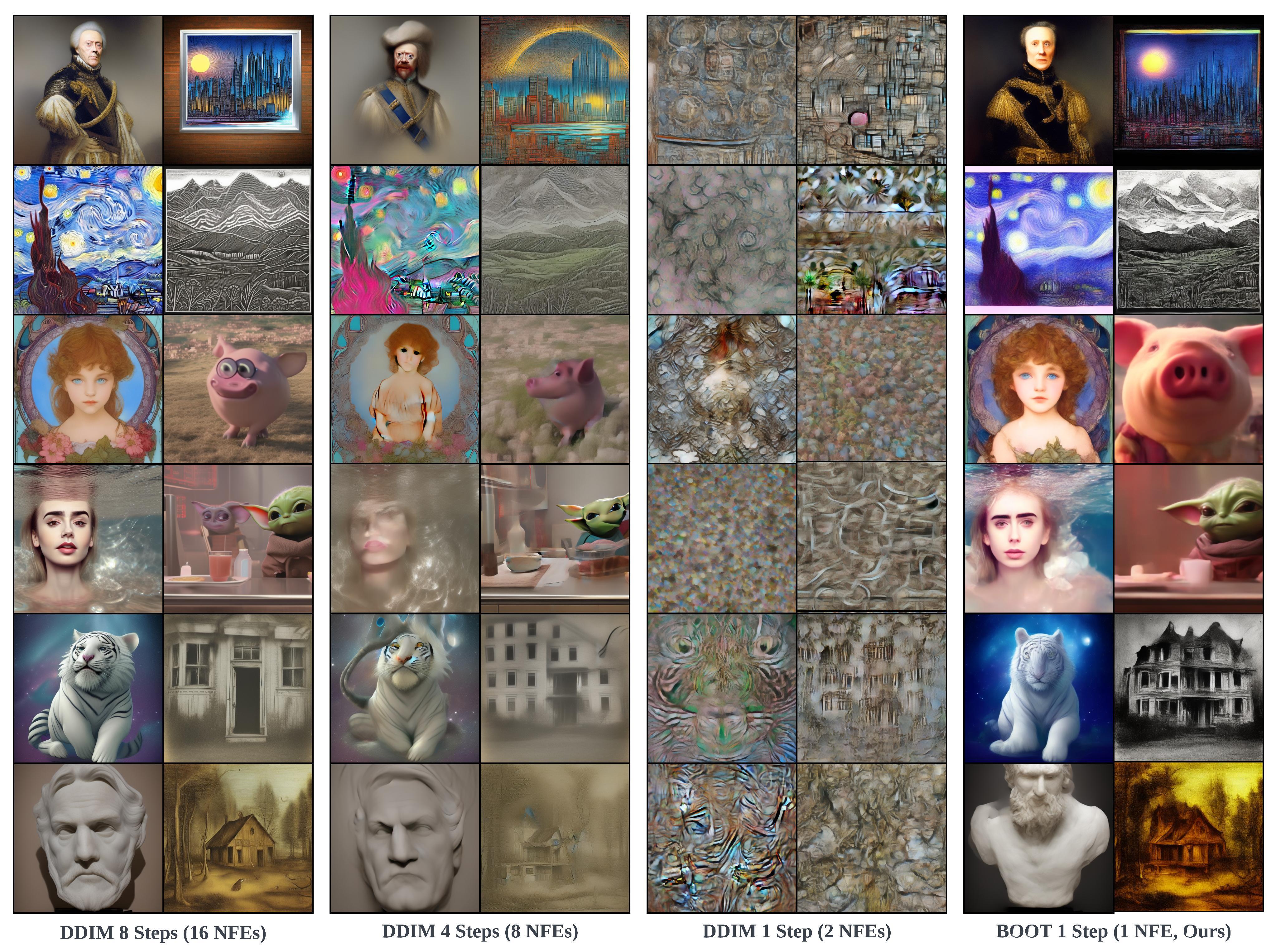}
    \caption{Uncurated samples of \{50, 10, 1\}  DDIM sampling steps and the proposed {\method} from SD2.1-base, given the same set of initial noise input and prompts sampled from \emph{diffusiondb}.}
    \vspace{-7pt}
    \label{fig:t2i}
\end{figure}

\vspace{-5pt}\paragraph{Evaluation Metrics}
For image generation, results are compared according to Fr\'echet Inception Distance (\citep[FID,][]{heusel2017gans}, lower is better), Precision (\citep[Prec.,][]{kynkaanniemi2019improved}, higher is better), and Recall (\citep[Rec.,][]{kynkaanniemi2019improved}, higher is better) over $50,000$ real samples from the corresponding datasets. 
For text-to-image tasks, we measure the zero-shot CLIP score~\citep{radford2021learning} for measuring the faithfulness of generation given $5000$ randomly sampled captions from COCO2017~\citep{Lin2014} validation set.
In addition, we also report the inference speed measured by \texttt{fps} with batch-size 1 on single A100 GPU.
\begin{figure}[t]
    \centering
    \begin{subfigure}{\textwidth}
    \centering
    \includegraphics[width=\linewidth]{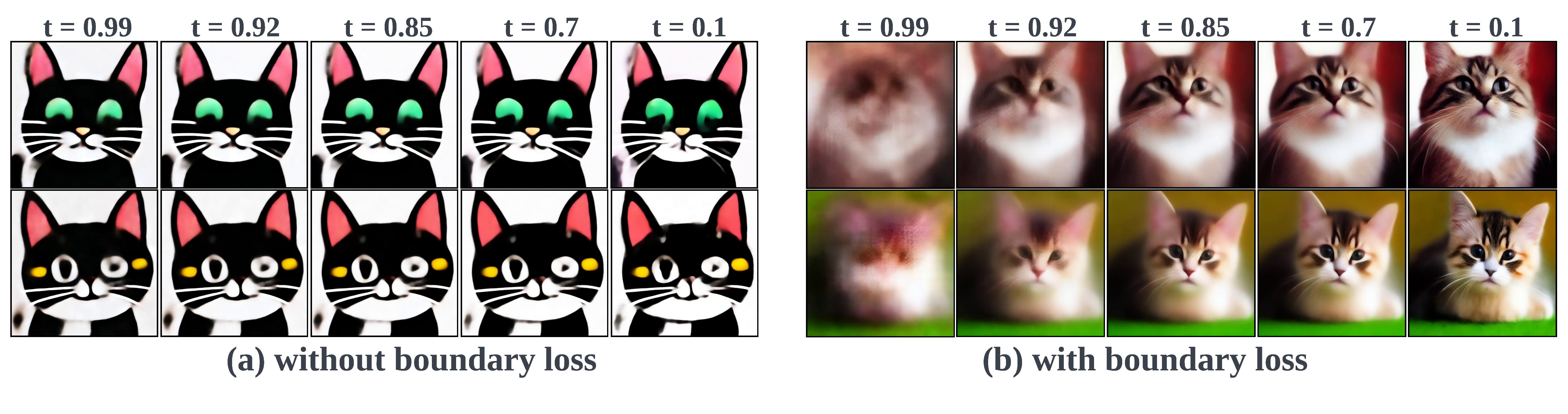}
     \vspace{-10pt}
   
    \end{subfigure}
    \begin{subfigure}{\textwidth}
    \centering
    \includegraphics[width=\linewidth]{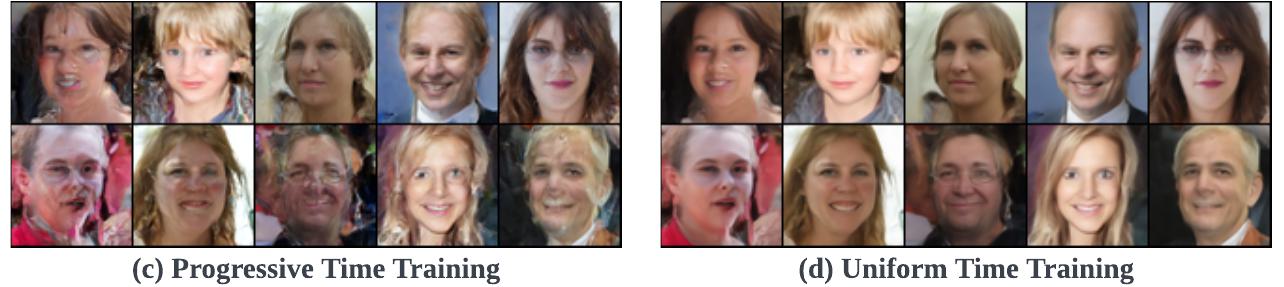}
  
    \end{subfigure}
    \caption{\label{fig:ablation}Ablation Study. (a) vs. (b): The additional boundary loss in \cref{sec.learning} alleviates the mode collapsing issue and prompts diversity in generation. (c) vs. (d): Uniform time training yields better generation compared with progressive time training. }
    \vspace{-4pt}
\end{figure}
\begin{figure}[t]
    \centering
    \includegraphics[width=\linewidth]{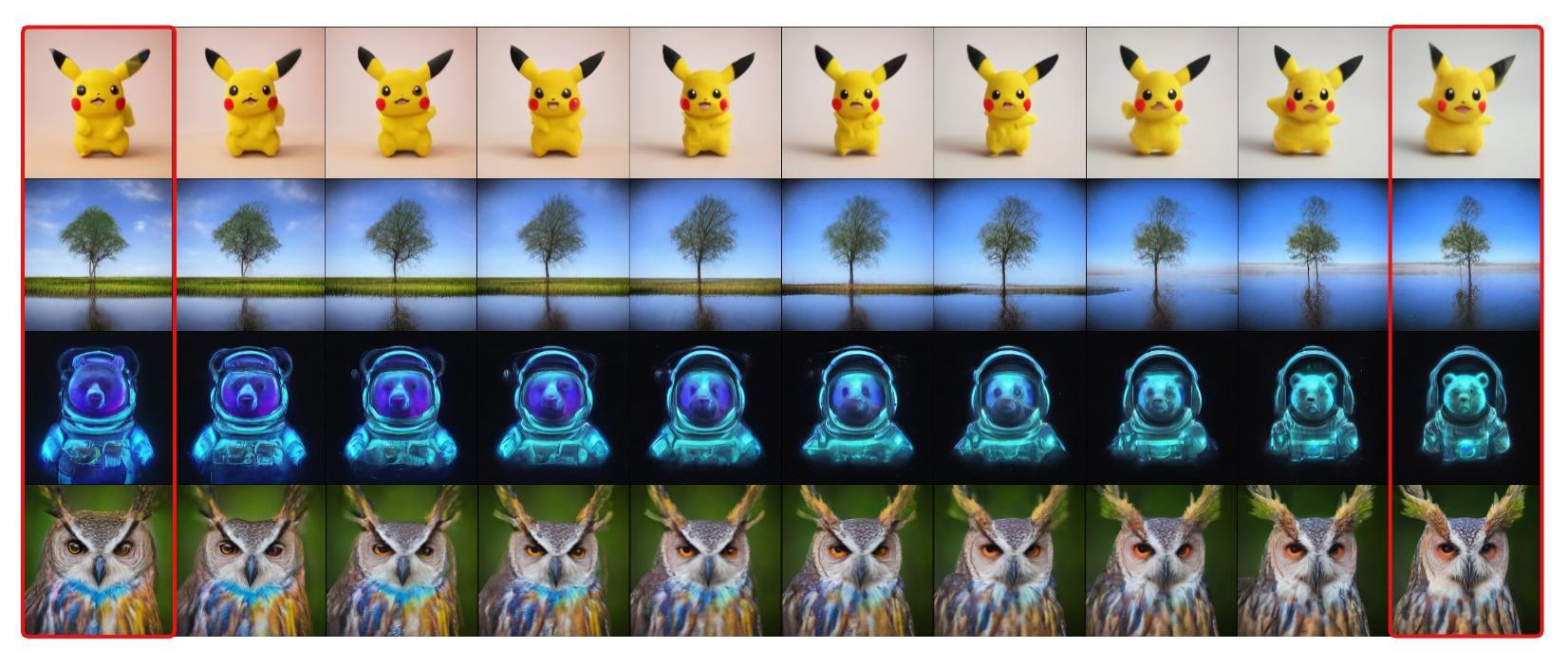}
    \vspace{-4pt}
    \caption{Latent space interpolation of the student model distilled from the IF teacher. 
    We randomly sample two noises to generate images (shown in red boxes) given the same text prompts, and then linearly interpolate the noises to synthesize images shown in the middle.
    }
    \vspace{-10pt}
    \label{fig:latent_interp}
\end{figure}

\subsection{Results}

\paragraph{Quantitative Results}
We first evaluate the proposed method on standard image generation benchmarks. The quantitative comparison with the standard diffusion inference methods like DDPM~\citep{ho2020denoising} and the deterministic DDIM~\citep{song2020denoising} are shown in \cref{tab:ffhq_imagenet}. 
Despite lagging behind the $50$-step DDIM inference, {\method} significantly improves the performance $1$-step inference, and achieves better performance against DDIM with around $10$ denoising steps, while maintaining $\times 10$ speed-up. Note that, the speed advantage doubles if the teacher employs guidance.

We also conduct quantitative evaluation on text-to-image tasks. Using the SD teacher, we obtain a CLIP-score of $0.254$ on COCO2017, a slight degradation compared to the $50$-step DDIM results ($0.262$), while it generates $2$ orders of magnitude faster, rendering real-time applications.

\vspace{-5pt}\paragraph{Visual Results}

We show the qualitative comparison in \cref{fig:image,fig:t2i} for image generation and text-to-image, respectively. For both cases, nav\"ie $1$-step inference fails completely, and the diffusion generally outputs grey and ill-structured images with fewer than $10$ NFEs. 
In contrast, {\method} is able to synthesize high-quality images that are visually close (\cref{fig:image}) or semantically similar (\cref{fig:t2i}) to teacher's results with much more steps.
Unlike the standard benchmarks, distilling text-to-image models (e.g., SD) typically leads to noticeably different generation from the original diffusion model, even starting with the same initial noise. We hypothesize it is a combined effect of highly complex underlying distribution and CFG.
We show more results including pixel-space models in the appendix.

\subsection{Analysis}
\paragraph{Importance of Boundary Condition}
The significance of incorporating the boundary loss is demonstrated in \cref{fig:ablation} (a) and (b). When using the same noise inputs, we compare the student outputs based on different target timesteps. As $\vy_\theta(\veps, t)$ tracks the signal-ODE output, it produces more averaged results as $t$ approaches 1. However, without proper boundary constraints, the student outputs exhibit consistent sharpness across timesteps, resulting in over-saturated and non-realistic images. This indicates a complete failure of the learned student model to capture the distribution of the teacher model, leading to severe mode collapse.

\vspace{-5pt}\paragraph{Progressive v.s. Uniform Time Training}
We also compare different training strategies in \cref{fig:ablation} (c) and (d). In contrast to the proposed approach of uniformly sampling $t$, one can potentially achieve additional efficiency with a fixed schedule that progressively decreases $t$ as training proceeds. This progressive training strategy seems reasonable considering that the student is always initialized from $t_{\max}$ and gradually learns to predict the clean signals (small $t$) during training. However, progressive training tends to introduce more artifacts (as observed in the visual comparison in \cref{fig:ablation}). We hypothesize that progressive training is more prone to accumulating irreversible errors.

\begin{figure}[t]
    \centering
    \includegraphics[width=0.97\linewidth]{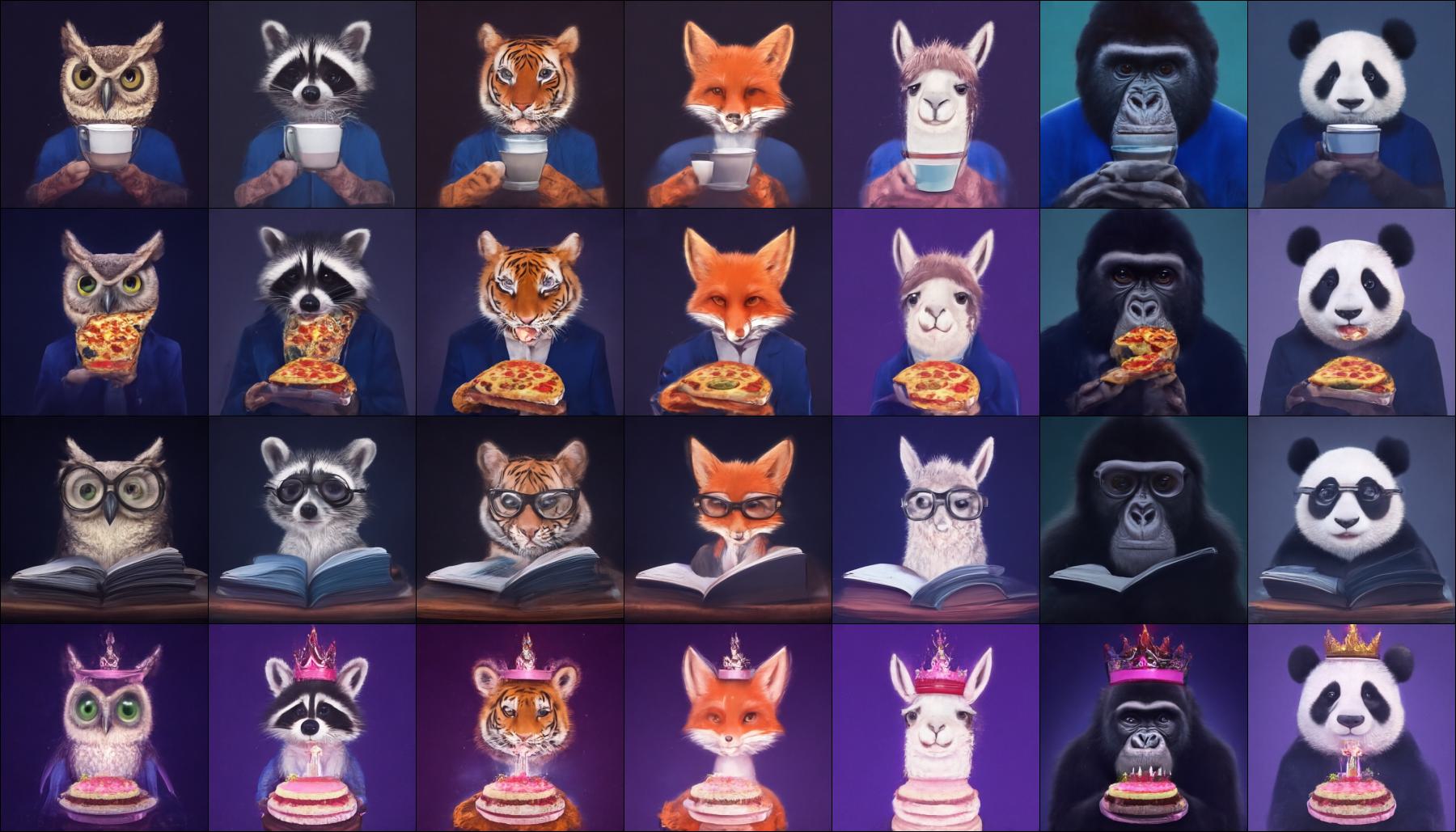}
    \caption{With fixed noise, we can perform controllable generation by swapping the keywords from the prompts. The prompts are chosen from the combination of \emph{portrait of a \{owl, raccoon, tiger, fox, llama, gorilla, panda\} wearing \{ a t-shirt, a jacket, glasses, a crown\} \{ drinking a latte, eating a pizza, reading a book, holding a cake\} cinematic, hdr.}
    All images are generated from the student distilled from IF teacher given the same noise input.
    }\vspace{-10pt}
    \label{fig:text_control2}
\end{figure}

\paragraph{Controllable Generation}
In \cref{fig:latent_interp}, we visualize the results of latent space interpolation, where the student model is distilled from the pretrained IF teacher. The smooth transition of the generated images demonstrates that the distilled student model has successfully learned a continuous and meaningful latent space. Additionally, in \cref{fig:text_control2}, we provide an example of text-controlled generation by fixing the noise input and only modifying the prompts. Similar to the original diffusion teacher model, the BOOT distilled student retains the ability of disentangled representation, enabling fine-grained control while maintaining consistent styles.

\section{Related Work}
\paragraph{Improving Efficiency of Diffusion Models} Speeding up inference of diffusion models is a broad area. Recent works and also our work ~\citep{luhman2021knowledge,salimans2022progressive,meng2022distillation,song2023consistency,berthelot2023tract} aim at reducing the number of diffusion model inference steps via distillation. 
Aside from distillation methods, other representative approaches include advanced ODE solvers~\citep{karras2022elucidating,lu2022dpm}, low-dimension space diffusion~\citep{rombach2021highresolution,vahdat2021score,jing2022subspace,gu2022f}, and improved diffusion targets~\citep{lipman2023flow,liu2022flow}. {\method} is orthogonal and complementary to these approaches, and can theoretically benefit from improvements made in all these aspects.

\vspace{-5pt}\paragraph{Knowledge Distillation for Generative Models}
Knowledge distillation~\citep{hinton2015distilling} has seen successful applications in learning efficient generative models, including model compression~\citep{kim2016sequence,aguinaldo2019compressing,fu2020autogan,hsieh2023distilling}
and non-autoregressive sequence generation~\citep{gu2017non,oord2018parallel,zhou2019understanding}. 
We believe that {\method} could inspire a new paradigm of distilling powerful generative models without requiring access to the training data.
\section{Discussion and Conclusion}
\paragraph{Limitations}
{\method} is a knowledge distillation algorithm, which by nature requires a pre-trained teacher model. Also by design, the sampling quality of BOOT is upper bounded by that of the teacher. Besides, {\method} may produce lower quality samples compared to other distillation methods~\citep{song2023consistency,berthelot2023tract} where ground-truth data are easy to use, which can potentially be remedied by combining methods. 

\paragraph{Future Work}
As future research, we aim to investigate the possibility of jointly training the teacher and the student models in a manner that incorporates the concept of diffusion into the distillation process. By making the diffusion process "distillation aware," we anticipate improved performance and more effective knowledge transfer. Furthermore, we find it intriguing to explore the training of a single-step diffusion model from scratch. This exploration could provide  insights into the applicability and benefits of {\method} in scenarios where a pre-trained model is not available.

\paragraph{Conclusion}
In summary, this paper introduced a novel technique \emph{{\method}} to distill diffusion models into single step. The method did not require the presence of any real or synthetic data by learning a time-conditioned student model with bootstrapping objectives. The proposed approach achieved comparable generation quality while being significantly faster than the diffusion teacher, and was also applicable to large-scale text-to-image generation, showcasing its versatility. 
\section*{Acknowledgement}
We thank Tianrong Chen, Miguel Angel Bautista, Navdeep Jaitly, Laurent Dinh, Shiwei Li, Samira Abnar, Etai Littwin 
for their critical suggestions and valuable feedback to this project. 

{\small
\bibliography{egbib}
\bibliographystyle{iclr2023_conference}
}

\begin{figure}[h!]
    \centering
    \vspace{-6pt}
    \includegraphics[width=0.96\linewidth]{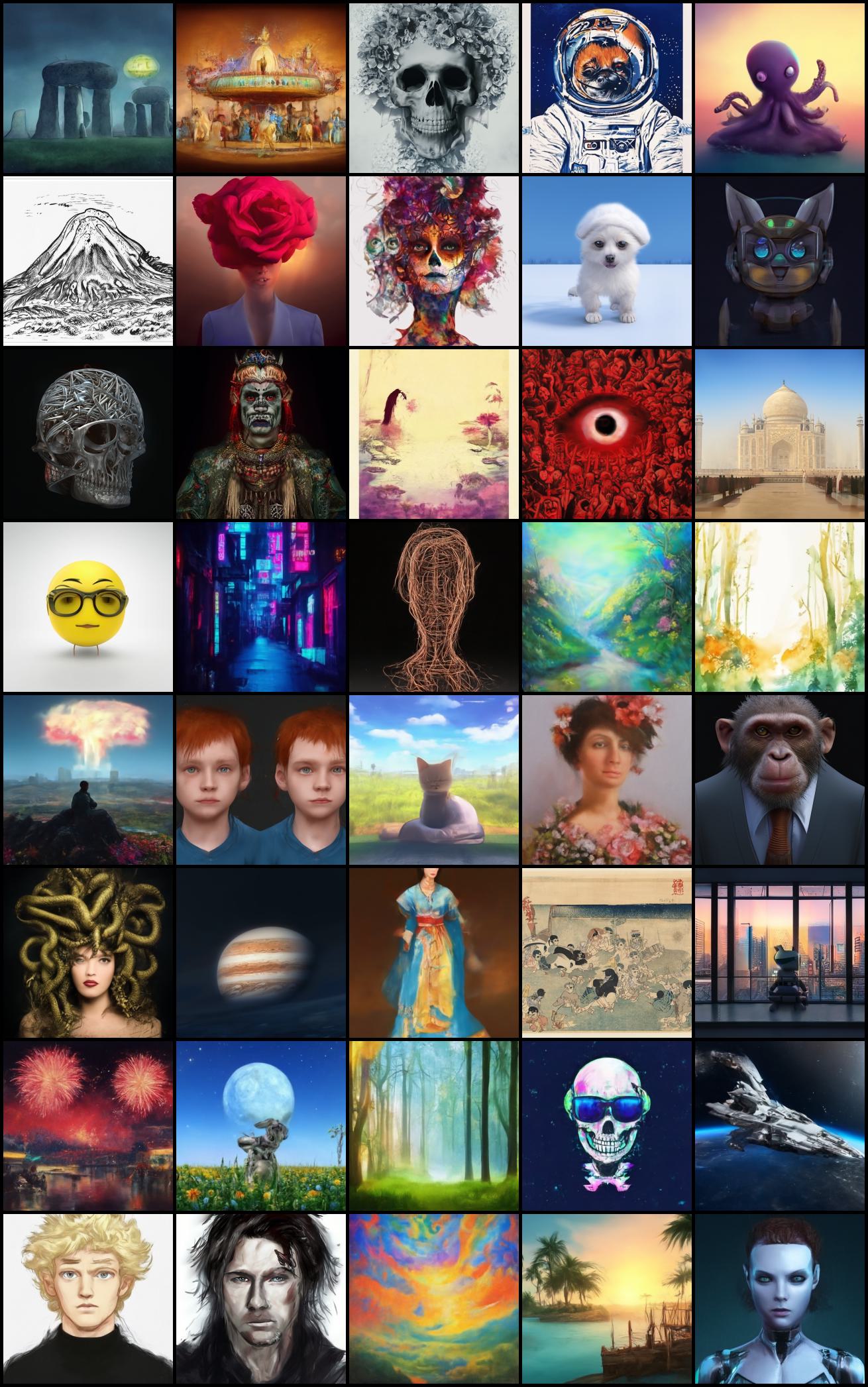}
    \caption{Curated samples of our distilled single-step model with prompts from \emph{diffusiondb}.}
    \label{fig:my_teaser2}
\end{figure}
\begin{figure}[h!]
    \centering
    \vspace{-6pt}
    \includegraphics[width=0.96\linewidth]{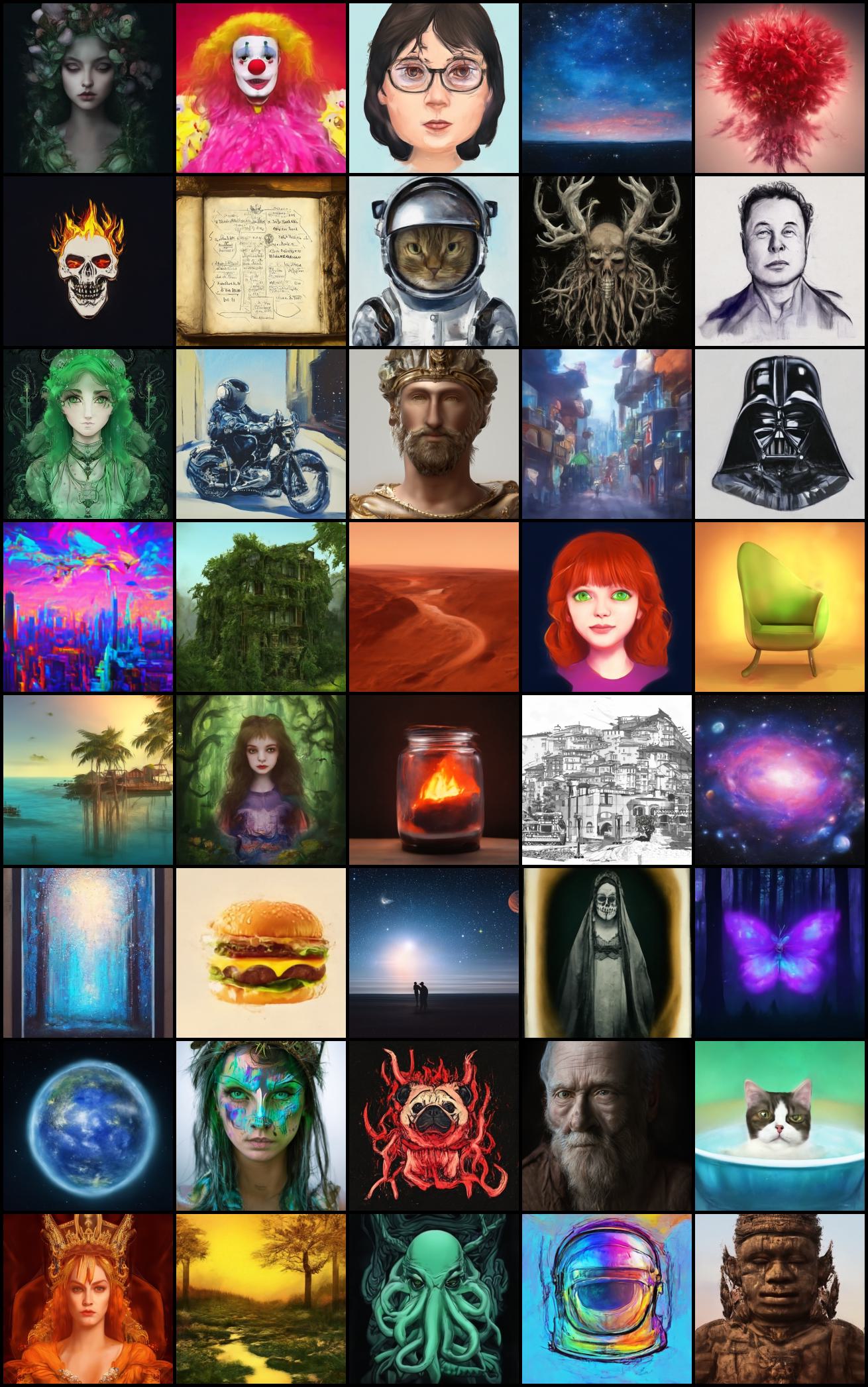}
    \caption{Curated samples of our distilled single-step model with prompts from \emph{diffusiondb}.}
    \label{fig:my_teaser3}
\end{figure}
\begin{figure}[h!]
    \centering
    \vspace{-6pt}
    \includegraphics[width=0.96\linewidth]{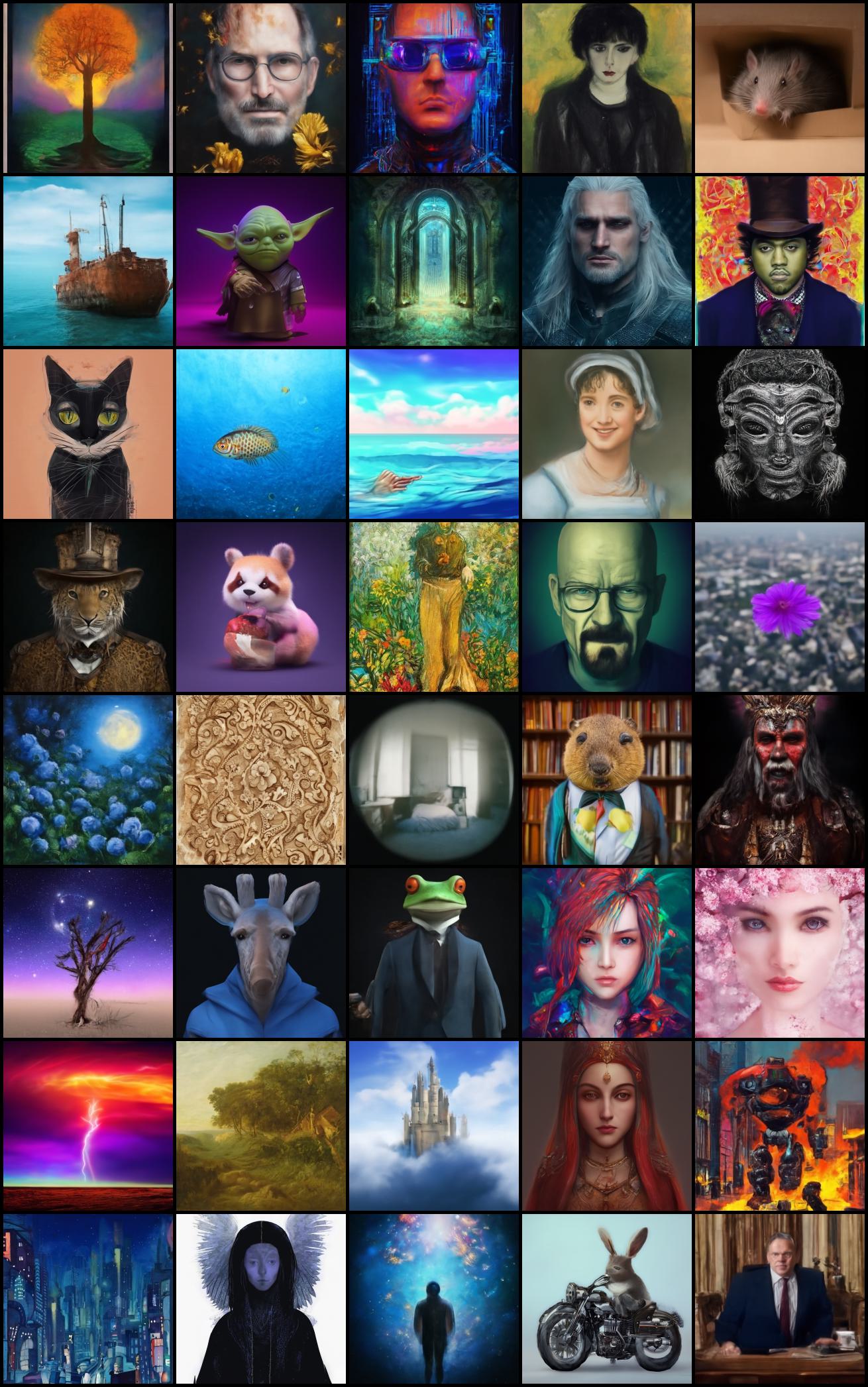}
    \caption{Curated samples of our distilled single-step model with prompts from \emph{diffusiondb}.}
    \label{fig:my_teaser1}
\end{figure}
\newpage
\appendix
\setcounter{tocdepth}{4}
\tableofcontents

\begin{appendices}
\newpage
\begin{center}
{\huge \bf Appendices}
\end{center}

\section{Algorithm Details}
\label{app:derivation}
\subsection{Notations}
In this paper, we use $\vf_\phi(\vx, t)$ to represent the diffusion model that denoises the noisy sample $\vx$ into its clean version, and we derive the DDIM sampler (\cref{eq.ddim}) following the definition of \citet{song2020denoising}: we deterministically synthesize $\vx_s$ based on the following update rule:
\begin{equation}
    \begin{split}
        \vx_s &= \texttt{ODE-Solver}(\vf_\phi, \veps, T\rightarrow s) \\
        &= \alpha_s \vf_\phi(\vx_t, t) + \sigma_s \left(\frac{\vx_t - \alpha_t\vf_\phi(\vx_t, t)}{\sigma_t}\right) \\
        &= \frac{\sigma_s}{\sigma_t}\vx_t + \left(\alpha_s - \alpha_t\frac{\sigma_s}{\sigma_t}\right)\vf_\phi(\vx_t, t) 
     \end{split}
     \label{eq.app_ddim}
\end{equation}
where $0 \leq s < t \leq T$. Here we use $\texttt{ODE-Solver}$ to represent the DDIM sampling from a random noise $\vx_T =\veps\sim \mathcal{N}(0, I)$, and iteratively obtain the sample at step $s$. In practice, we can generalize to higher-order ODE-solvers for better efficiency.

For distillation, we define the student model with $\vg_\theta(\veps, t)$ which approximates $\vx_t$ along the diffusion trajectory above. 
To avoid directly predicting the noisy samples $\vx_t$ with neural networks, we re-parameterize $\vg_\theta(\veps, t) = \alpha_t \vy_\theta(\veps, t) + \sigma_t\veps$ where the noise part is constant throughout $t$ except the scale factor $\sigma_t$. 
In this way, the learning goal $\vy_\theta(\veps, t)$ is to predict a new variable $\vy_t$: the ``signal'' part of the original variable $\vy_t = (\vx_t - \sigma_t\veps) / \alpha_t$. 

\subsection{Derivation of Signal-ODE}
Based on the definition of $\vy_t = (\vx_t - \sigma_t\veps)/\alpha_t$,
we can derive the following equations from \cref{eq.app_ddim}:
\begin{equation}
    \begin{split}
        \vx_s &= \frac{\sigma_s}{\sigma_t}\vx_t + \left(\alpha_s - \alpha_t\frac{\sigma_s}{\sigma_t}\right)\vf_\phi(\vx_t, t)\\
         \Rightarrow \alpha_s \vy_s + \sigma_s\veps &= \frac{\sigma_s}{\sigma_t}\left(\alpha_t \vy_t + \sigma_t\veps\right) + \left(\alpha_s - \alpha_t\frac{\sigma_s}{\sigma_t}\right)\vf_\phi(\vx_t, t) \\
        \Rightarrow \alpha_s \vy_s + \cancel{\sigma_s\veps}  &= \alpha_t\frac{\sigma_s}{\sigma_t} \vy_t + \cancel{\sigma_s\veps} + \left(\alpha_s - \alpha_t\frac{\sigma_s}{\sigma_t}\right)\vf_\phi({\vx}_t, t)  \\
         \Rightarrow \vy_s   &= \frac{\alpha_t\sigma_s}{\sigma_t\alpha_s} \vy_t + \left(1 - \frac{\alpha_t\sigma_s}{\sigma_t\alpha_s}\right)\vf_\phi({\vx}_t, t) \\
         &= \left(1 - e^{\lambda_s - \lambda_t}\right) \vf_\phi({\vx}_t, t) + e^{\lambda_s - \lambda_t}\vy_t,
    \end{split}
    \label{eq.app_update}
\end{equation}
where we use the auxiliary variable $\lambda_t = -\log(\alpha_t / \sigma_t)$ for simplifying the equations.
As mentioned in \cref{sec.signal}, we can further obtain the continuous form of \cref{eq.app_update} by assigning $t - s \rightarrow 0$. That is, \cref{eq.app_update} is equivalent to that shown in the following:
\begin{equation}
    \begin{split}
    \vy_s &= \left(1 - e^{\lambda_s - \lambda_t}\right) \vf_\phi({\vx}_t, t) + e^{\lambda_s - \lambda_t}\vy_t\\
        \Rightarrow \vy_t - \vy_s &= -\left(1 - e^{\lambda_s - \lambda_t}\right) \left( \vf_\phi({\vx}_t, t) - \vy_t\right)\\
        \Rightarrow \frac{\vy_t - \vy_s}{t - s} &= -\frac{e^{\lambda_t} - e^{\lambda_s}}{t - s}\cdot e^{-\lambda_t}\left( \vf_\phi({\vx}_t, t) - \vy_t\right) \\
        \Rightarrow \frac{\diff\vy_t}{\diff t} &= - \cancel{e^{\lambda_t}}\cdot \lambda'_t \cdot \cancel{e^{-\lambda_t}} \left( \vf_\phi({\vx}_t, t) - \vy_t\right) 
    \end{split}
    \label{eq.app_ode}
\end{equation}
where $\lambda'_t = {\diff\lambda_t}/{\diff t}$. Given a fixed noise input $\veps$, \cref{eq.app_ode} defines an ODE over $\vy_\theta$ w.r.t $t$, which we call \emph{Signal-ODE}, as both sides of the equation only operate in ``low-frequency'' signal space.
\begin{algorithm}[t]
  \caption{Distillation using {\method} for Conditional Diffusion Models.\label{alg:distillation}}
  \begin{algorithmic}[1]
    \Require{pretrained diffusion model $\vf_\phi$, initial student parameter from the teacher $\theta\leftarrow\phi$, 
    step size $\delta$,
    learning rate $\eta$, CFG weight $w$, 
    context dataset $\mathcal{D}$, 
    negative condition $\vn=\emptyset$,
    $t_{\min}, t_{\max}$, $\beta$.}
    \While{not converged}
        \State Sample noise input $\veps \sim \mathcal{N}(0, I)$
        \State Sample context input $\vc\sim \mathcal{D}$
        \State Sample $t\sim (t_{\min}, t_{\max}), s = \min\left(t - \delta, t_{\min})\right)$
        \State Compute noise schedule $\alpha_t, \sigma_t, \alpha_s, \sigma_s$
        \State Compute $\lambda'_t\approx(1 - \dfrac{\alpha_t\sigma_s}{\sigma_t\alpha_s})/\delta$
        \State Generate the model predictions: \\
        \;\;\;\;\;\;\;\;\;\;\;\; $\vy_t = \vy_\theta(\veps, t, \vc)$, \;\; $\vy_s = \vy_\theta(\veps, s, \vc)$, \;\; $\vy_{t_{\max}} = \vy_\theta(\veps, t_{\max}, \vc)$
        \State Generate the noisy sample $\hat{\vx}_t = \alpha_t\vy_t + \sigma_t\veps$
        \State Compute the denoised target: \\
        \;\;\;\;\;\;\;\;\;\;\;\;  $\tilde{\vf}_t = \vf_\phi(\hat{\vx}_t, t, \vn) + w\cdot\left(
          \vf_\phi(\hat{\vx}_t, t, \vc) - \vf_\phi(\hat{\vx}_t, t, \vn)
          \right)$\\
        \;\;\;\;\;\;\;\;\;\;\;\;  $\tilde{\vf}_{t_{\max}} = \vf_\phi(\veps, t_{\max}, \vn) + w\cdot\left(
          \vf_\phi(\veps, t_{\max}, \vc) - \vf_\phi(\veps, t_{\max}, \vn)
          \right)$  
        \State Compute the bootstrapping loss
        $\mathcal{L}^\textrm{BS}_\theta = \dfrac{1}{(\delta\lambda'_t)^2}\|\vy_s - \texttt{SG}(\vy_t + \delta\lambda'_t (\tilde{\vf}_t - \vy_t))\|^2_2$
        \State Compute the boundary loss 
        $\mathcal{L}^\textrm{BC}_\theta = \|\vy_{t_{\max}} - \tilde{\vf}_{t_{\max}}\|^2_2$
        \State Update model parameters $\theta \gets \theta - \eta \cdot \nabla_\theta \left(\mathcal{L}^\textrm{BS}_\theta + \beta \mathcal{L}^\textrm{BC}_\theta\right)$
    \EndWhile
    
    \State \textbf{return} Trained model parameters $\theta$
  \end{algorithmic}
\end{algorithm}
\subsection{Bootstrapping Objectives}
The bootstrapping objectives in \cref{eq.training} can be easily derived by taking the finite difference of \cref{eq.cm_loss}. Here we use $\vy_\theta(\veps, t)$ to estimate $\vy_t$, and use $\hat{\vx}_t$ to represent the noisy image obtained from $\vy_\theta(\veps, t)$.
\begin{equation}
    \begin{split}
        \mathcal{L}_\theta &= \mathbb{E}_{\veps, t}
     \left[\tilde{\omega}_t
     \left|\left|
        \frac{\diff\vy_\theta(\veps, t)}{\diff t} + \lambda'_t\cdot\left(\vf_\phi(\hat{\vx}_t, t) -\vy_\theta(\veps, t) \right)
     \right|\right|^2_2\right] \\
      &\approx \mathbb{E}_{\veps, t}
     \left[\tilde{\omega}_t
     \| \frac{\vy_\theta(\veps, s) - \vy_\theta(\veps, t)}{\delta} - 
        \lambda'_t \left(\vf_\phi(\hat{\vx}_t, t) -\vy_\theta(\veps, t) \right)\|^2_2
     \right] \\
     &= \mathbb{E}_{\veps, t}
     \left[\frac{\tilde{\omega}_t}{\delta^2}
     \| \vy_\theta(\veps, s) - \left[\vy_\theta(\veps, t) + \delta\lambda'_t \left(\vf_\phi(\hat{\vx}_t, t) -\vy_\theta(\veps, t) \right)\right]\|^2_2
     \right] \\
     &= \mathbb{E}_{\veps, t}
     \left[\frac{\tilde{\omega}_t}{\delta^2}
     \| \vy_\theta(\veps, s) 
        - \hat{\vy}_\theta(\veps, s)
     \|^2_2
     \right],
    \end{split}
\end{equation}
where $s = t-\delta$, and $\hat{\vy}_\theta(\veps, s)$ is the approximated target.
$\tilde{\omega}_t$ is the additional weight, where by default $\tilde{\omega}_t = 1$.
 To stabilize training, a stop-gradient operation $\texttt{SG}(.)$ is typically included:
\begin{equation}
     \mathcal{L}_\theta = \mathbb{E}_{\veps, t}
     \left[\frac{\tilde{\omega}_t}{\delta^2}
     \| \vy_\theta(\veps, s) 
        - \texttt{SG}(\hat{\vy}_\theta(\veps, s))
     \|^2_2
     \right].
\end{equation}
In our experiments, we also find that it helps use $\tilde{\omega}_t =1/{\lambda'^2_t}$ for text-to-image generation.

We can take advantage of higher-order solvers for a more accurate target that reduces the discretization error. For example, one can use Heun's method~\citep{ascher1998computer} to first calculate the intermediate value $\tilde{\vy}_\theta(\veps, s)$, and then the final approximation $\hat{\vy}_\theta(\veps, s)$:
\begin{equation}
    \begin{split}
        \tilde{\vy}_\theta(\veps, s) &= \vy_\theta(\veps, t) + \delta\lambda'_t \left(\vf_\phi(\hat{\vx}_t, t) -\vy_\theta(\veps, t) \right), \; \;
        \tilde{\vx}_s = \alpha_s \tilde{\vy}_\theta(\veps, s) + \sigma_s\veps \\
        \hat{\vy}_\theta(\veps, s) &= \vy_\theta(\veps, t) + \frac{\delta\lambda'_t}{2}\left[
            \left(\vf_\phi(\hat{\vx}_t, t) -\vy_\theta(\veps, t) \right) + 
            \left(\vf_\phi(\tilde{\vx}_s, s) -\tilde{\vy}_\theta(\veps, s) \right)
        \right].
    \end{split}
\end{equation}
Using Heun's method essentially doubles the evaluations of the teacher model during training, while the add-on overheads are manageable as we stop the gradients to the teacher model.

\subsection{Training Algorithm}
We summarize the training algorithm of {\method} in \cref{alg:distillation}, where by default we assume conditional diffusion model with classifier-free guidance and DDIM solver. 
Here, for simplicity, we write $\lambda'_t\approx(1 - \frac{\alpha_t\sigma_s}{\sigma_t\alpha_s})/\delta$.
For unconditional models, we can simply remove the context sampling part.

\section{Connections to Existing Literature}
\subsection{Physics Informed Neural Networks (PINNs)}
Physics-Informed Neural Networks~\citep[PINNs,][]{raissi2019physics} are powerful approaches that combine the strengths of neural networks and physical laws to solve ODEs. Unlike traditional numerical methods, which rely on discretization and iterative solvers, PINNs employ machine learning techniques to approximate the solution of ODEs. The key idea behind PINNs is to incorporate physics-based constraints directly into the training process of neural networks. By embedding the governing equations and available boundary or initial conditions as loss terms, PINNs can effectively learn the underlying physics while simultaneously discovering the solution. This ability makes PINNs highly versatile in solving a wide range of ODEs, including those arising in fluid dynamics, solid mechanics, and other scientific domains. Moreover, PINNs offer several advantages, such as automatic discovery of spatio-temporal patterns and the ability to handle noisy or incomplete data.

Although motivated from different perspectives, {\method} shares similarities with PINNs at a high level, as both aim to learn ODE/PDE solvers directly through neural networks. In the domain of PINNs, solving ODEs can also be simplified into two objectives: the differential equation (DE) loss (\cref{eq.de}) and the boundary condition (BC) loss (\cref{eq.boundary}). The major difference lies in the focus of the two approaches. PINNs primarily focus on learning complex ODEs/PDEs for single problems, where neural networks serve as universal approximators to address the discretization challenges faced by traditional solvers. Moreover, the data space in PINNs is relatively low-dimensional. In contrast, {\method} aims to learn single-step generative models capable of synthesizing data in high-dimensional spaces (e.g., millions of pixels) from random noise inputs and conditions (e.g., labels, prompts). To the best of our knowledge, no existing work has applied similar methods in generative modeling. Additionally, while standard PINNs typically compute derivatives (\cref{eq.de}) directly using auto-differentiation, in this paper, we employ the finite difference method and propose a bootstrapping-based algorithm.

\subsection{Consistency Models / TRACT}
The most related previous works to our research are Consistency Models~\citep{song2023consistency} and concurrently TRACT~\citep{berthelot2023tract}, which propose bootstrapping-style algorithms for distilling diffusion models. These approaches map an intermediate noisy training example at time step $t$ to the teacher's $t$-step denoising outputs using the DDIM inference procedure. The training target for the student is constructed by running the teacher model with one step, followed by the self-teacher with $t - 1$ steps.
As illustrated in \cref{fig:compare_with_cm}, {\method} takes a different approach to bootstrapping. It starts from the Gaussian noise prior and directly maps it to an intermediate step $t$ in one shot. This change has significant modeling implications, as it does not require any training data and can achieve data-free distillation, a capability that none of the prior works possess.

\subsection{Single-step Generative Models}
{\method} is also related to other single-step generative models, including VAEs~\citep{kingma2013auto} and GANs~\citep{goodfellow2014generative}, which aim to synthesize data in a single forward pass. However, {\method} does not require an encoder network like VAEs. Thanks to the power of the underlying diffusion model, {\method} can produce higher-contrast and more realistic samples. In comparison to GANs,
{\method} does not require a discriminator or critic network. Furthermore, the distillation process of {\method} enables better-controlled exploration of the text-image joint space, which is explored by the pretrained diffusion models, resulting in more coherent and realistic samples in text-guided generation. Additionally, {\method} is more stable to learn compared to GANs, which are challenging to train due to the adversarial nature of maintaining a balance between the generator and discriminator networks.

\section{Additional Experimental Settings}
\label{app:details}
\subsection{Datasets}
While the proposed method is data-free, we list the additional dataset information that used to train our teacher diffusion models:

\textbf{FFHQ} (\url{https://github.com/NVlabs/ffhq-dataset}) contains 70k images of real human faces in resolution of $1024\times 1024$. In most of our experiments, we resize the images to a low resolution at $64\times 64$ for early-stage benchmarking.

\textbf{LSUN} (\url{https://www.yf.io/p/lsun}) is a collection of large-scale image dataset containing 10 scenes and 20 object categories. Following previous works~\citep{song2023consistency}, we choose the category Bedroom ($3$M images), and train an unconditional diffusion teacher. All images are resized to $256\times 256$ with center-crop. We use LSUN to validate the ability of learning in relative high-resolution scenarios.

\textbf{ImageNet-1K} (\url{https://image-net.org/download.php}) contains $1.28$M images across $1000$ classes. We directly merge all the training images with class labels and train a class-conditioned diffusion teacher. All images are resized to $64\times 64$ with center-crop. To support test-time classifier-free guidance, the teacher model is trained with $0.2$ unconditional probability.

As we do not need to train our own teacher models for text-to-image experiments, no additional text-image pairs are required in this paper. However, our distillation still requires the text conditions for querying the teacher diffusion. To better capture and generalize the real user preference of such diffusion models, we choose to adopt the collected prompt datasets:

\textbf{DiffusionDB} (\url{https://poloclub.github.io/diffusiondb/}) contains $14$M images generated by Stable Diffusion using prompts and hyperparameters specified by users. For the purpose of our experiments, we only keep the text prompts and discard all model-generated images as well as meta-data and hyperparameters so that they can be used for different teacher models. We use the same prompts for both latent and pixel space models.

\subsection{Text-to-Image Teachers}
We directly choose the recently open-sourced large-scale diffusion models as our teacher models. More specifically, we looked into the following models:

\textbf{StableDiffusion (SD)} (\url{https://github.com/Stability-AI/stablediffusion})  is an open-source text-to-image latent diffusion model~\citep{rombach2021highresolution} conditioned on the penultimate text embeddings of a CLIP ViT-H/14~\citep{radford2021learning} text encoder. 
Different standard diffusion models, SD performs diffusion purely in the latent space.  
In this work, we use the checkpoint of \textbf{SD v2.1-Base} (\url{https://huggingface.co/stabilityai/stable-diffusion-2-1-base}) as our teacher which first generates in $64\times 64$ latent space, and then directly upscaled to $512\times 512$ resolution with the pre-trained VAE decoder. The teacher model was trained on subsets of LAION-5B~\citep{schuhmann2022laion} with noise prediction objective.

\begin{table}[t]
    \centering
    \begin{adjustbox}{max width=\linewidth}
        \begin{tabular}{l|ccc|ccc}
            \toprule
            & \multicolumn{3}{c|}{Image Generation} & \multicolumn{3}{c}{Text-to-Image} \\
            Hyperparameter & \multicolumn{1}{c}{FFHQ} & \multicolumn{1}{c}{LSUN} & \multicolumn{1}{c|}{ImageNet} &
            SD-Base & IF-I-L & IF-II-M
            \\
            \midrule
            Architecture \\
            \;\; Denosing resolution &
            $64\times 64$ & $256\times 256$ & $64\times 64$ & $64\times 64$ & $64\times 64$ & $256\times 256$
            \\ 
            \;\; Base channels & 128 & 128 & 192 & \\
            \;\; Multipliers & 1,2,3,4 & 1,1,2,2,4,4 & 1,2,3,4  &\\
            \;\; \# of Resblocks & 1 & 1 & 2 & \\
            \;\; Attention resolutions & 8,16 & 8,16 & 8,16 &  & -- Default --&\\
            \;\; Noise schedule & cosine & cosine & cosine & \\
            \;\; Model Prediction & signal & signal & signal &\\
            \;\; Text Encoder & - &- & - & CLIP & T5 & T5 \\
            \midrule
            Training \\
            \;\; Loss weighting & uniform & uniform & uniform & $\lambda'^{-2}_t$ & $\lambda'^{-2}_t$ & $\lambda'^{-2}_t$
            \\
            \;\; Bootstrapping step size & 0.04 & 0.04 & 0.04 & 0.01 & 0.04 & 0.04 \\
            \;\; CFG weight & - & - & $1\sim 5$ & $7.5$ & $7.0$ & $4.0$
            \\
            \;\; Learning rate & 1e-4 & 1e-4 & 3e-4 & 2e-5 & 2e-5 & 2e-5\\
            \;\; Batch size & 128 & 128 & 1024 & 64 & 64 & 32 \\
            \;\; EMA decay rate & 0.9999 & 0.9999 & 0.9999 & 0.9999 & 0.9999 & 0.9999 \\
            \;\; Training iterations & 500k & 500k & 300k & 500k & 500k & 100k\\
            \bottomrule
        \end{tabular}
    \end{adjustbox}
    \vspace{5pt}
    \caption{Hyperparameters used for training {\method}. The CFG weights for text-to-image models are determined based on the default value of the open-source codebase.}\label{tab:hyperparameters}
\end{table}

\textbf{DeepFloyd IF (IF)} (\url{https://github.com/deep-floyd/IF})  is a recently open-source text-to-image model with a high degree of photorealism and language understanding. IF is a modular composed of a frozen text encoder and three cascaded pixel diffusion modules, similar to Imagen~\citep{saharia2022photorealistic}: a base model that generates $64\times 64$ image based on text prompt and two super-resolution models ($256\times 256, 1024\times 1024$). 
All stages of the model utilize a frozen text encoder based on the T5~\citep{raffel2020exploring} to extract text embeddings, which are then fed into a UNet architecture enhanced with cross-attention and attention pooling. Models were trained on 1.2B text-image pairs (based on LAION~\citep{schuhmann2022laion} and few additional internal datasets) with noise prediction objective.
In this paper, we conduct experiments on the first two resolutions ($64\times 64, 256\times 256$) with the checkpoints of \textbf{IF-I-L-v1.0} (\url{https://huggingface.co/DeepFloyd/IF-I-L-v1.0}) and \textbf{IF-II-M-v1.0} (\url{https://huggingface.co/DeepFloyd/IF-II-M-v1.0}).

\subsection{Model Architectures}
We follow the standard U-Net architecture~\citep{nichol2021improved} for image generation benchmarks and adopt the hyperparameters similar in f-DM~\citep{gu2022f}. For text-to-image applications, we keep the default architecture setups from the teacher models unchanged.
As mentioned in the main paper, we initialize the weights of the student models directly from the pretrained checkpoints and use \emph{zero} initialization for the newly added modules, such as target time and CFG weight embeddings. 
We include additional architecture details in the \cref{tab:hyperparameters}.

\subsection{Training Details}
All models for all the tasks are trained on the same resources of $8$ NVIDIA A100 GPUs for $500$K updates. Training roughly takes $3\sim7$ days to converge depending on the model sizes. We train all our models with the AdamW~\citep{loshchilov2017decoupled} optimizer, with no learning rate decay or warm-up, and no weight decay. Standard EMA to the weights is also applied for student models.
Since our methods are data-free, there is no additional overhead on data storage and loading except for the text prompts, which are much smaller and can be efficiently loaded into memory.

Learning the boundary loss requires additional NFEs during each training step. In practice,  we apply the boundary loss less frequently (e.g., computing the boundary condition every $4$ iterations and setting the loss to be $0$ otherwise) to improve the overall training efficiency.
Note that distilling from the class-conditioned / text-to-image teachers requires multiple forward passes due to CFG, which relatively slows down the training compared to unconditional models.

Distilling from the DeepFloyd IF teacher requires learning from two stages. In this paper, we can easily achieve that by first distilling the first-stage model into single-step with {\method}, and then distilling the upscaler model based on the output of the first-stage student. Following the original paper~\citep{saharia2022photorealistic}, noise augmentation is also applied on the first-stage output where we set the noise-level as $250$~\footnote{\url{https://github.com/huggingface/diffusers/blob/main/src/diffusers/pipelines/deepfloyd_if/pipeline_if_superresolution.py\#L715}}.
For more training hyperparameters, please refer to \cref{tab:hyperparameters}.

\section{Additional Samples from {\method}}
\begin{figure}[t]
    \centering
    \includegraphics[width=\linewidth]{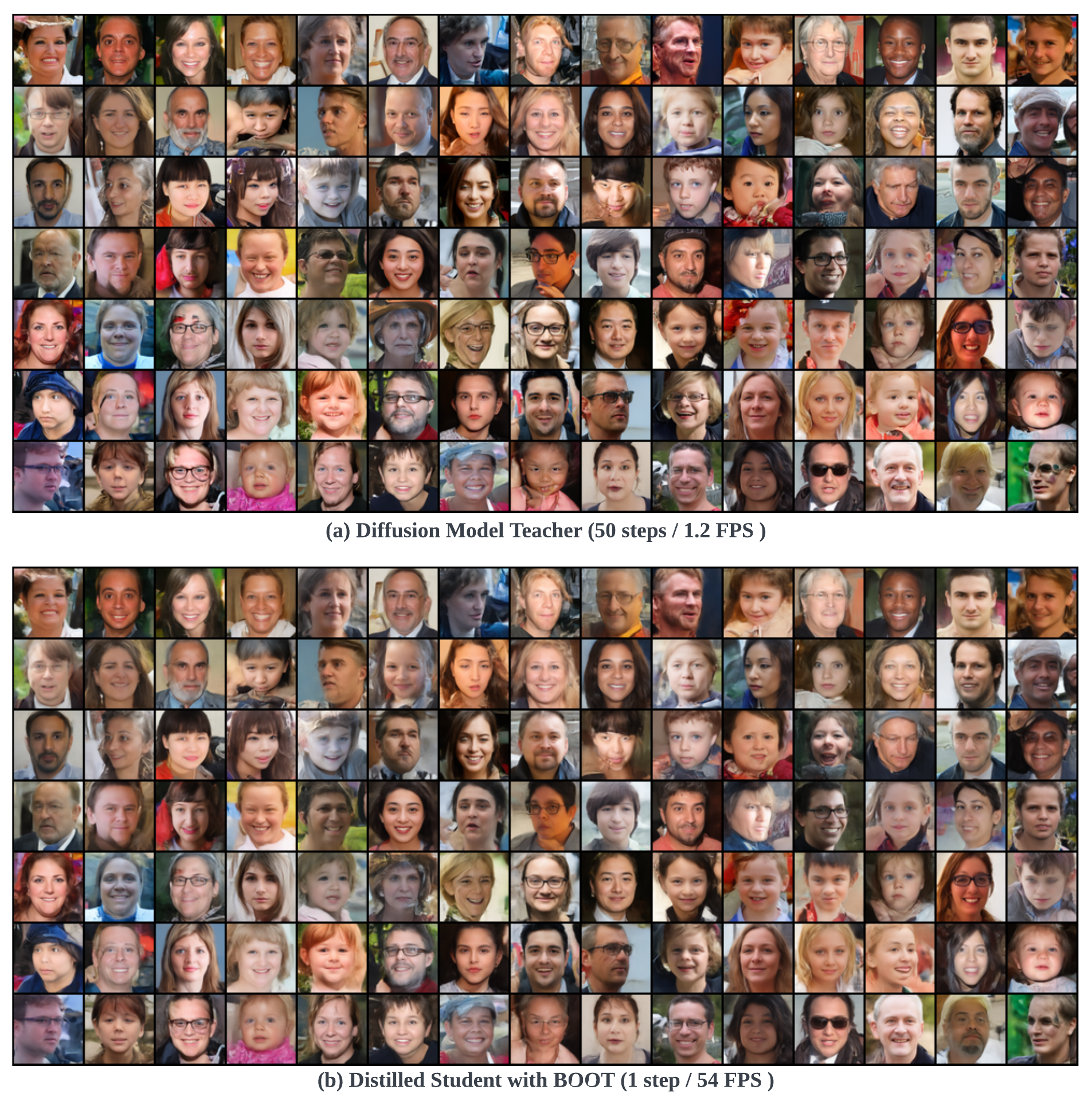}
    \caption{Uncurated samples from FFHQ $64\times 64$. All corresponding samples use the same initial noise for the DDIM teacher and the single-step {\method} student.}
    \label{fig:app_ffhq}
\end{figure}

\begin{figure}[t]
    \centering
    \includegraphics[width=\linewidth]{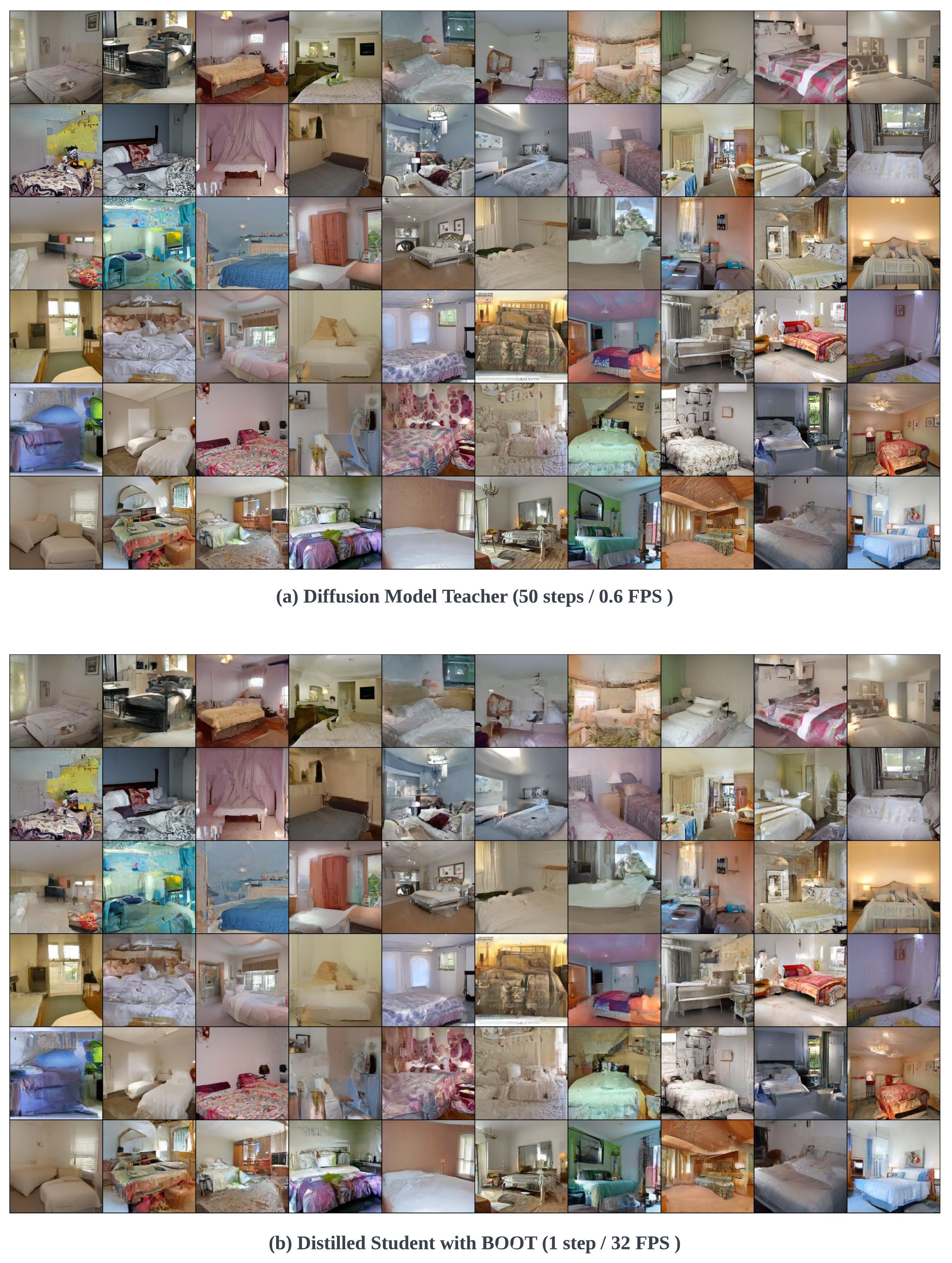}
    \caption{Uncurated samples from LSUN Bedroom $256\times 256$. All corresponding samples use the same initial noise for the DDIM teacher and the single-step {\method} student.}
    \label{fig:app_lsun}
\end{figure}

\begin{figure}[t]
    \centering
    \includegraphics[width=\linewidth]{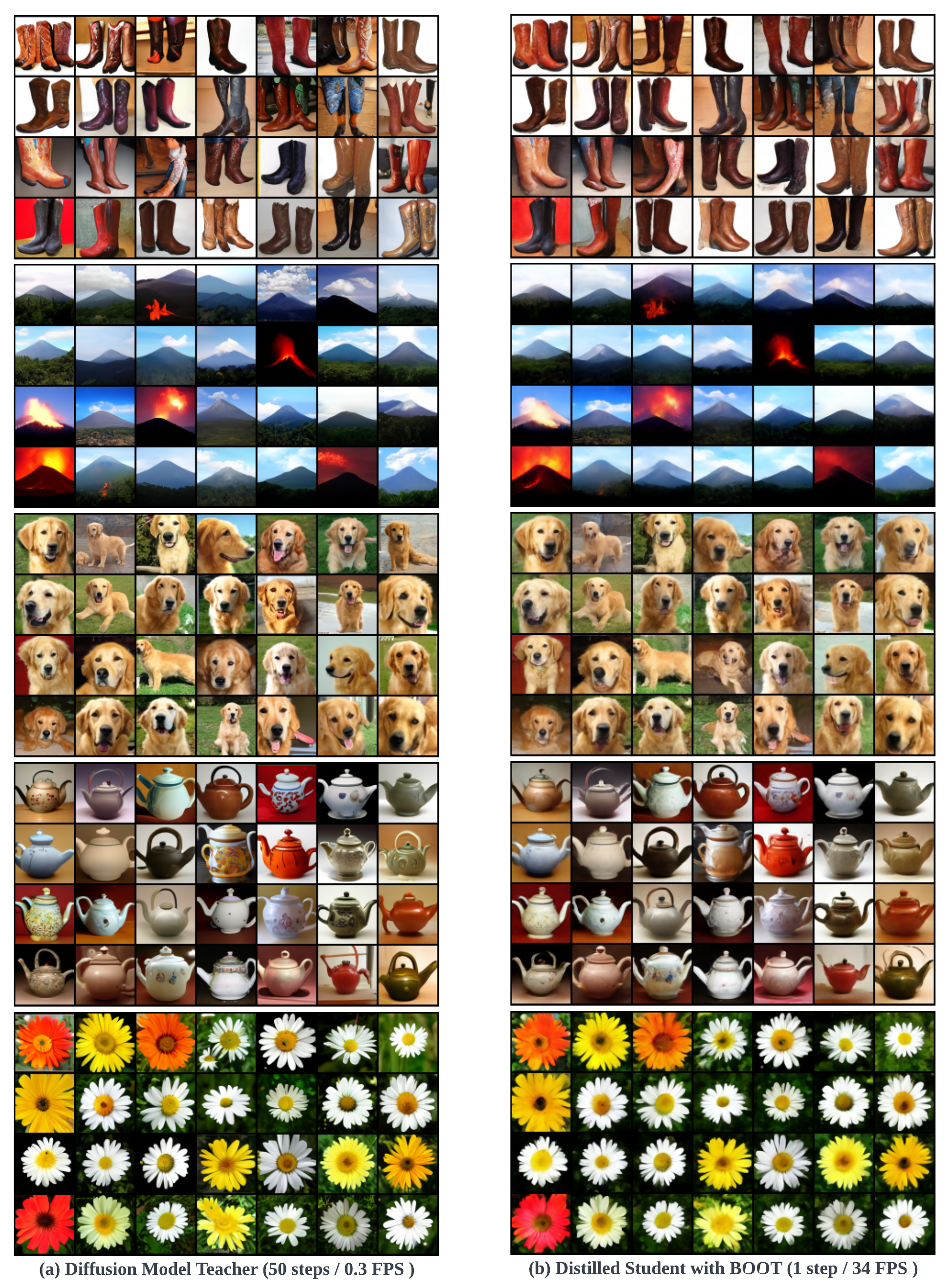}
    \caption{Uncurated class-conditioned samples from ImageNet $64\times 64$. All corresponding samples use the same initial noise for the DDIM teacher and the single-step {\method} student. Classes from top to bottom: \emph{cowboy boot, volcano, golden retriever, teapot, daisy}. The diffusion model uses CFG with $w=3$, and our student model conditions on the same weight.}
    \label{fig:app_imagenet}
\end{figure}

\begin{figure}[t]
    \centering
    \includegraphics[width=\linewidth]{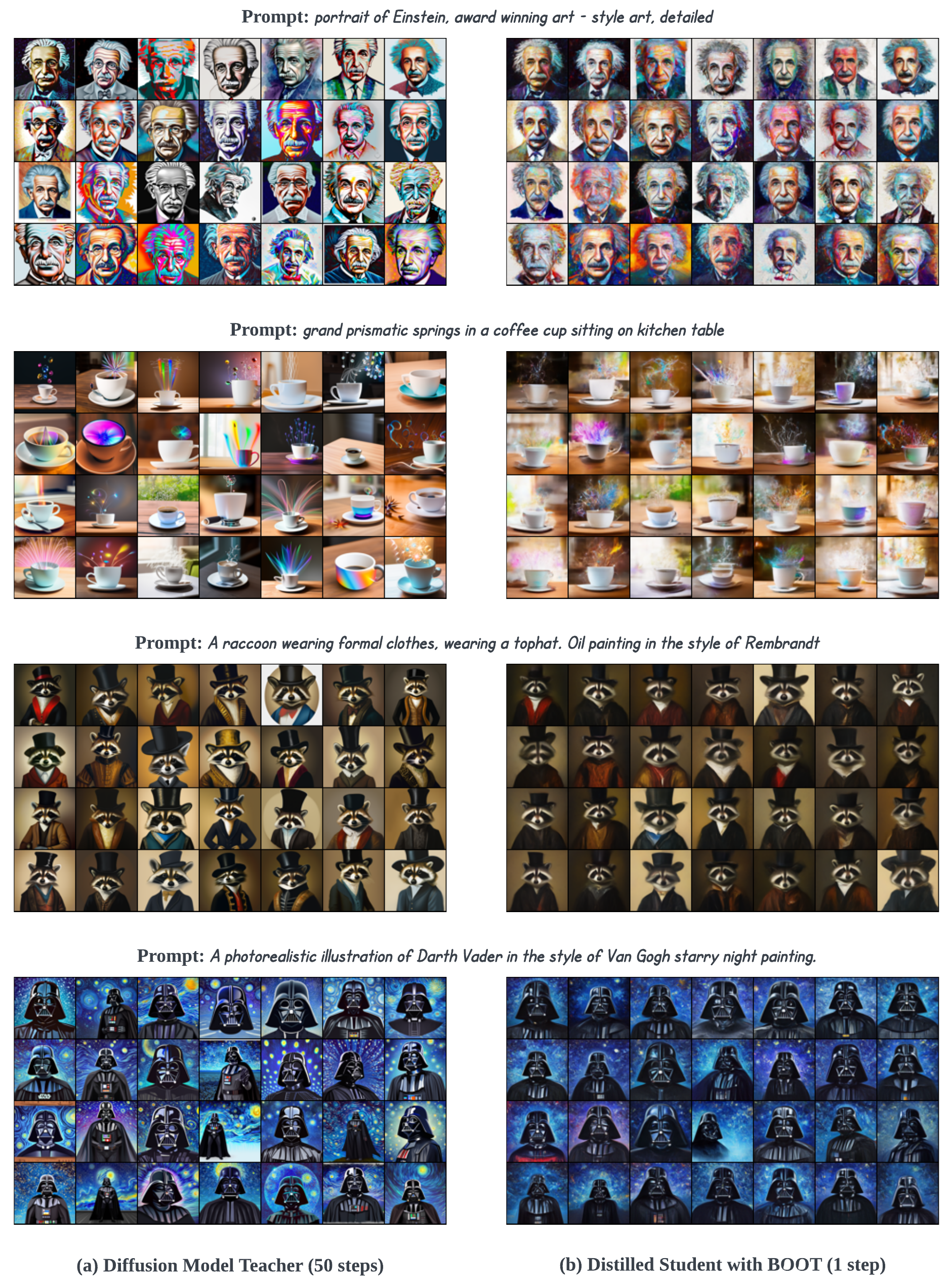}
    \caption{Uncurated text-conditioned image generation distilled from DeepFloyd IF (the first stage model, images are at $64\times 64$). All corresponding samples use the same initial noise for the DDIM teacher and the single-step {\method} student. The specific prompts are shown above the images.}
    \label{fig:app_if}
\end{figure}

\begin{figure}[t]
    \centering
    \includegraphics[width=\linewidth]{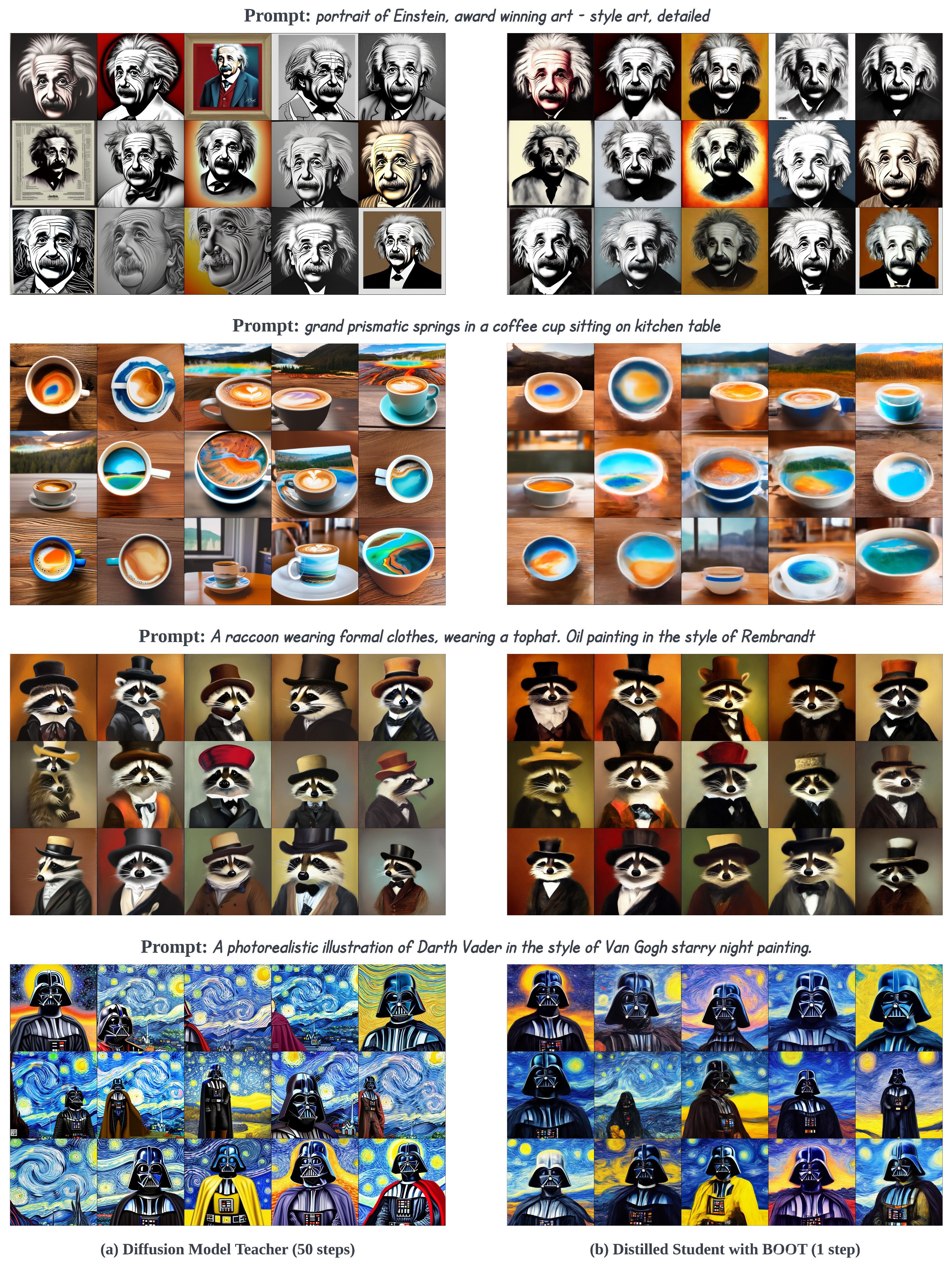}
    \caption{Uncurated text-conditioned image generation distilled from StableDiffusion (latent diffusion in $64\times 64$, images are upscaled to $512\times 512$ with the pre-trained VAE decoder). All corresponding samples use the same initial noise for the DDIM teacher and the single-step {\method} student. We use the same prompts as in \cref{fig:app_if} for better comparison.}
    \label{fig:app_sd}
\end{figure}

\begin{figure}[t]
    \centering
    \includegraphics[width=\linewidth]{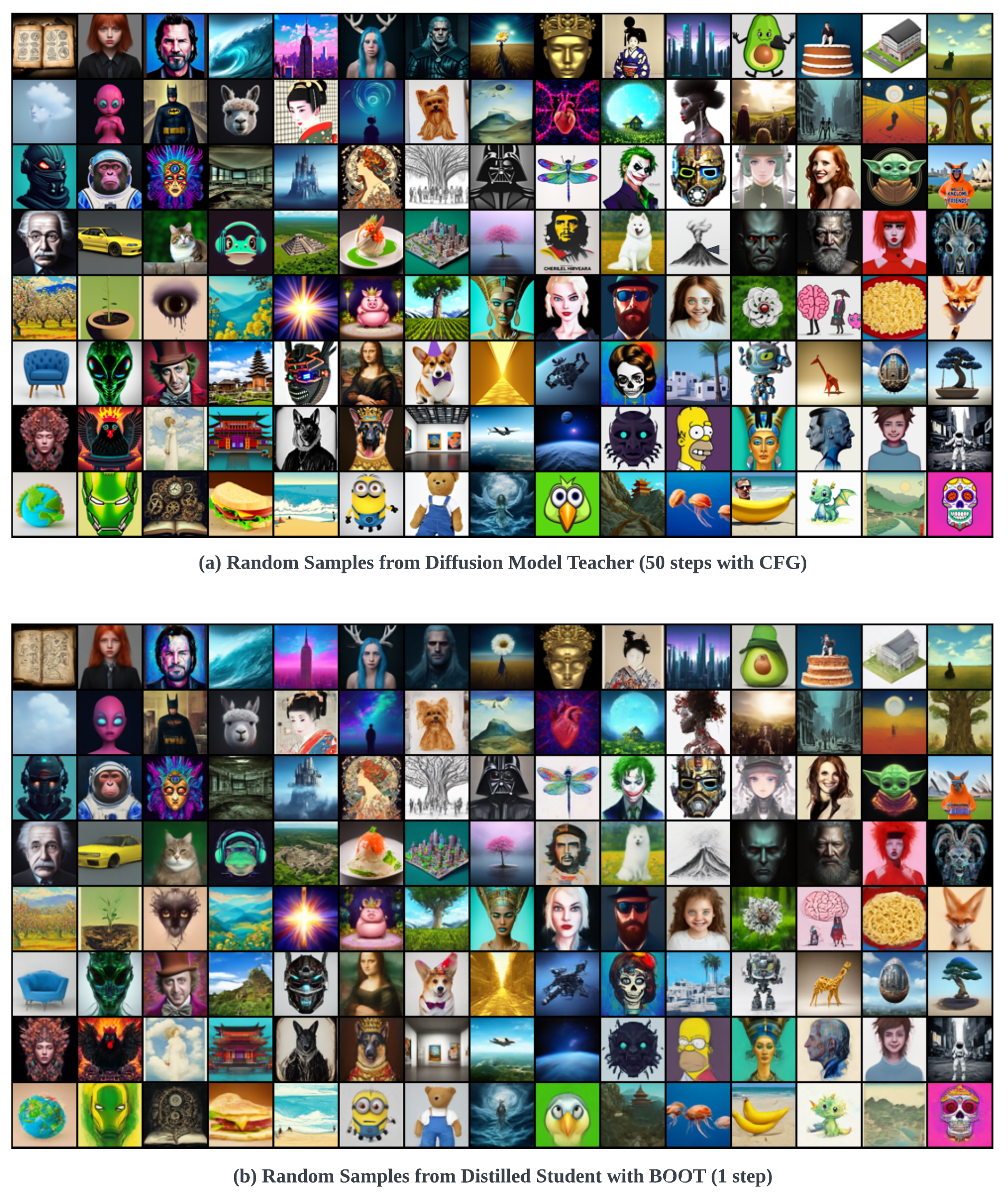}
    \caption{Uncurated text-conditioned image generation distilled from DeepFloyd IF (the first stage model, images are at $64\times 64$) given sampled text prompts from \textit{diffusiondb}~\citep{wangDiffusionDBLargescalePrompt2022} randomly. All corresponding samples use the same initial noise for the DDIM teacher and the single-step student. 
    Besides, we also show curated examples from the two-stage distilled model at $256\times 256$  in \cref{fig:my_teaser}.
    }
    \label{fig:app_if_random}
\end{figure}

\begin{figure}[t]
    \centering
    \includegraphics[width=\linewidth]{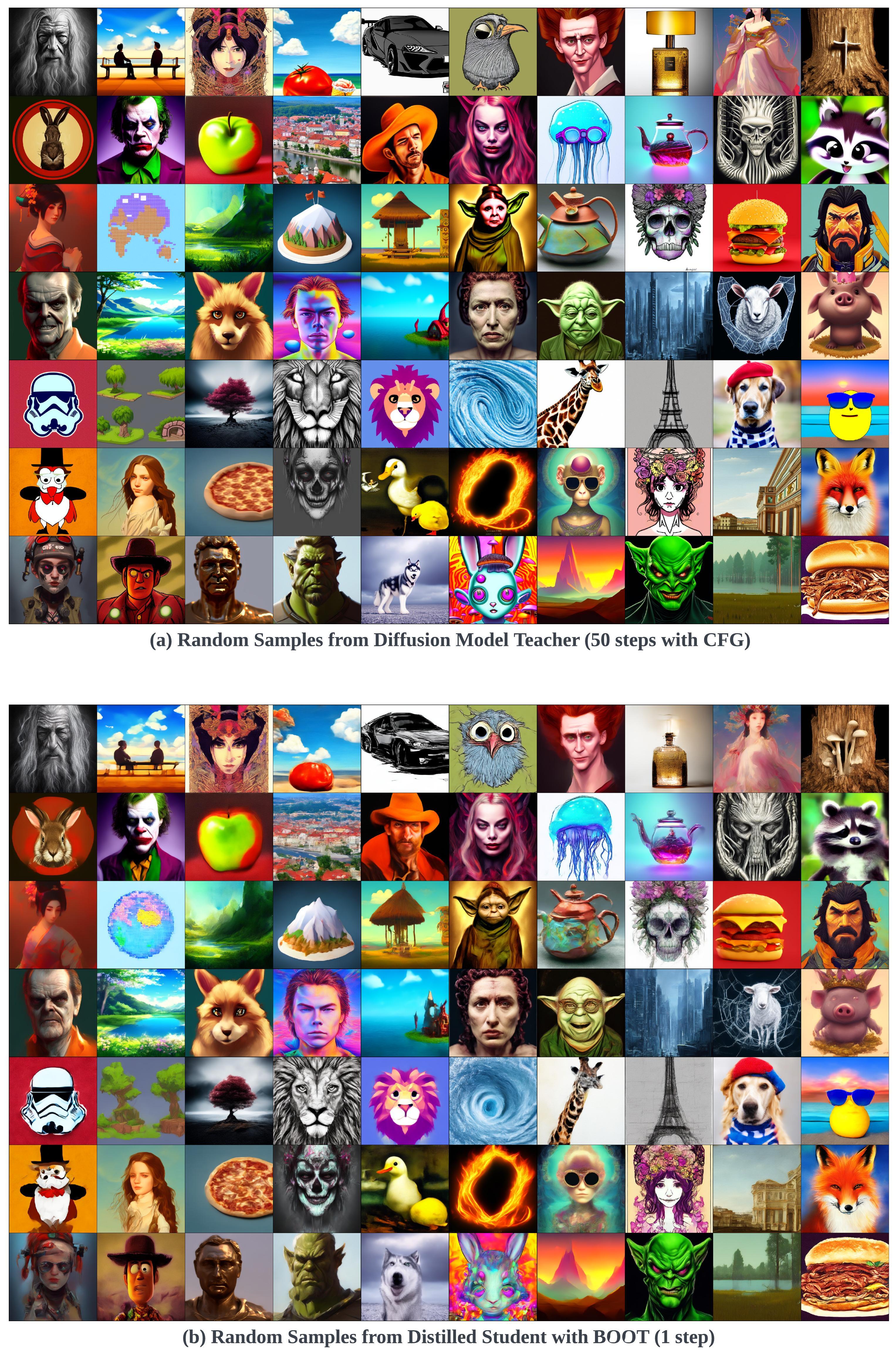}
    \caption{Uncurated text-conditioned image generation distilled from StableDiffusion (latent diffusion in $64\times 64$, images are upscaled to $512\times 512$ with the pre-trained VAE decoder) given sampled text prompts from \textit{diffusiondb}~\citep{wangDiffusionDBLargescalePrompt2022} randomly. All corresponding samples use the same initial noise for the DDIM teacher and the single-step student. }
    \label{fig:app_sd_random}
\end{figure}

Finally, we provide additional qualitative comparisons for the unconditional models of FFHQ $64\times 64$ (\cref{fig:app_ffhq}),  LSUN $256\times 256$ (\cref{fig:app_lsun}), the class-conditional model of ImageNet $64\times 64$ (\cref{fig:app_imagenet}), and comparisons for text-to-image generation based on DeepFloyd-IF ($64\times 64$ in \cref{fig:app_if,fig:app_if_random},  $256\times 256$ in \cref{fig:my_teaser,fig:my_teaser1,fig:my_teaser2,fig:my_teaser3}) and StableDiffusion ($512\times 512$ in \cref{fig:app_sd,fig:app_sd_random}).

\begin{figure}[t]
    \centering
    \includegraphics[width=\linewidth]{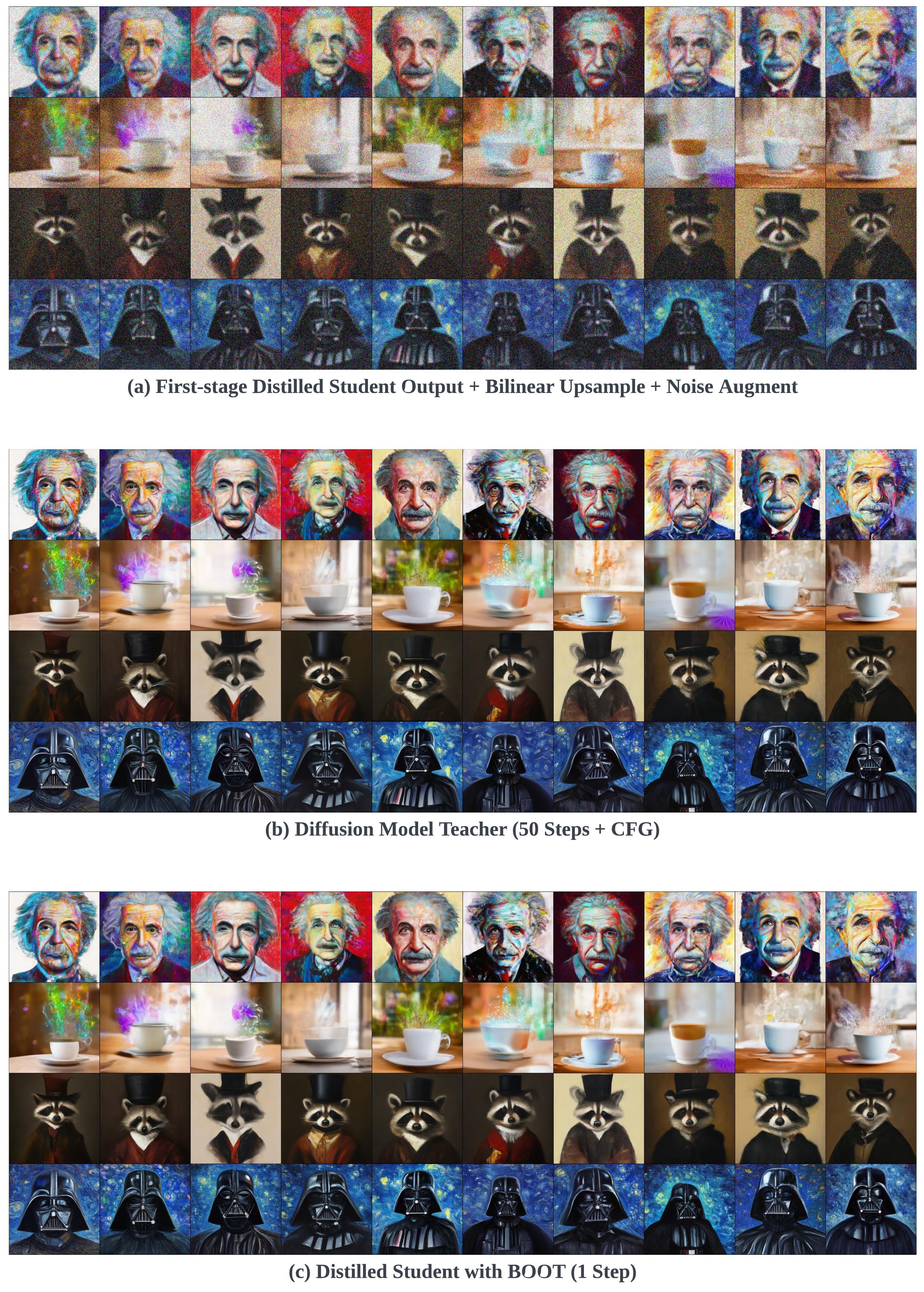}
    \caption{Given the $64\times 64$ outputs from \cref{fig:app_if}, we also show comparison for the second-stage models which upscale the images to $256\times 256$. All corresponding samples use the same initial noise for the DDIM teacher and the single-step {\method} student.}
    \label{fig:app_if_upsample}
\end{figure}

\end{appendices}

\end{document}